%% file: icml.tex
\documentclass{article}

\usepackage{microtype}
\usepackage{graphicx}
\usepackage{subfigure}
\usepackage{booktabs} 
\usepackage{hyperref}

\usepackage[accepted]{icml2024}
\usepackage{times}
\usepackage{latexsym}
\usepackage[T1]{fontenc}
\usepackage[utf8]{inputenc}
\usepackage{microtype}
\usepackage{xcolor}
\usepackage{inconsolata}
\usepackage{colortbl}
\usepackage{paralist}
\usepackage{amsmath}
\usepackage{amssymb}
\usepackage{bbm}
\usepackage{threeparttable}
\usepackage{multirow}
\usepackage{graphicx}
\usepackage{booktabs}
\usepackage{arydshln}
\usepackage{longtable}
\usepackage{footnote}
\usepackage{tablefootnote}
\usepackage{afterpage}
\usepackage{chngcntr}
\usepackage{listings}

\usepackage{amsmath}
\usepackage{amssymb}
\usepackage{mathtools}
\usepackage{amsthm}
\usepackage{siunitx} 
\usepackage[capitalize,noabbrev]{cleveref}

\theoremstyle{plain}

\theoremstyle{definition}

\theoremstyle{remark}

\usepackage[textsize=tiny]{todonotes}
\newcommand{\BenchmarkName}{TravelPlanner}

\icmltitlerunning{\BenchmarkName: A Benchmark for Real-World Planning with Language Agents}

\begin{document}

\twocolumn[
\icmltitle{\BenchmarkName: A Benchmark for Real-World Planning with Language Agents}

\icmlsetsymbol{equal}{*}

\begin{icmlauthorlist}
\icmlauthor{Jian Xie\textsuperscript{\rm $\spadesuit$}}{equal}\footnotetext{Equal Contribution}
\icmlauthor{Kai Zhang\textsuperscript{\rm $\clubsuit$}}{equal}
\icmlauthor{Jiangjie Chen\textsuperscript{\rm $\spadesuit$}}{}
\icmlauthor{Tinghui Zhu\textsuperscript{\rm $\spadesuit$}}{}
\icmlauthor{Renze Lou\textsuperscript{\rm $\heartsuit$}}{} \\ 
\icmlauthor{Yuandong Tian\textsuperscript{\rm $\diamondsuit$}}{}
\icmlauthor{Yanghua Xiao\textsuperscript{\rm $\spadesuit$}}{}
\icmlauthor{Yu Su\textsuperscript{\rm $\clubsuit$}}{}
\\
\textsuperscript{\rm $\spadesuit$}Fudan University
\textsuperscript{\rm $\clubsuit$}The Ohio State University  \\
\textsuperscript{\rm $\heartsuit$}The Pennsylvania State University
\textsuperscript{\rm $\diamondsuit$}Meta AI
\\{\small \texttt{jianxie22@m.fudan.edu.cn, shawyh@fudan.edu.cn, \{zhang.13253, su.809\}@osu.edu}}
\\
\url{https://osu-nlp-group.github.io/TravelPlanner}
\end{icmlauthorlist}

\icmlkeywords{Machine Learning, ICML}

\vskip 0.3in
]



\printAffiliationsAndNotice{\icmlEqualContribution} 

\begin{abstract}
\input{010abstract}
\end{abstract}

\section{Introduction}
\input{020introduction}

\section{Related Work}
\input{030relatedWork}

\section{\BenchmarkName}
\input{040benchmark}

\section{Experiments}

\input{050experiments}

\section{In-Depth Analysis}
\input{060analysis}

\section{Conclusion}
\input{070conclusion}

\section{Impact Statements}
\input{080impact}

\bibliography{icml}
\bibliographystyle{icml2024}

\newpage
\appendix
\clearpage
\onecolumn
\counterwithout*{footnote}{section}
\section*{Appendices}
\label{sec:appendix}
\input{090appendix}

\end{document}

%% file: 010abstract.tex
Planning has been part of the core pursuit for artificial intelligence since its conception, but earlier AI agents mostly focused on constrained settings because many of the cognitive substrates necessary for human-level planning have been lacking.
Recently, language agents powered by large language models (LLMs) have shown interesting capabilities such as tool use and reasoning.
\textit{Are these language agents capable of planning in more complex settings that are out of the reach of prior AI agents?}
To advance this investigation, we propose \BenchmarkName, a new planning benchmark that focuses on travel planning, a common real-world planning scenario.
It provides a rich sandbox environment, various tools for accessing nearly four million data records, and \num{1225} meticulously curated planning intents and reference plans.
Comprehensive evaluations show that the current language agents are not yet capable of handling such complex planning tasks---even GPT-4 only achieves a success rate of \num{0.6}\%.
Language agents struggle to stay on task, use the right tools to collect information, or keep track of multiple constraints.
However, we note that the mere possibility for language agents to tackle such a complex problem is in itself non-trivial progress.
\BenchmarkName{} provides a challenging yet meaningful testbed for future language agents. 

%% file: 020introduction.tex
\begin{figure*}
    \centering
    \includegraphics[width=.95\linewidth]{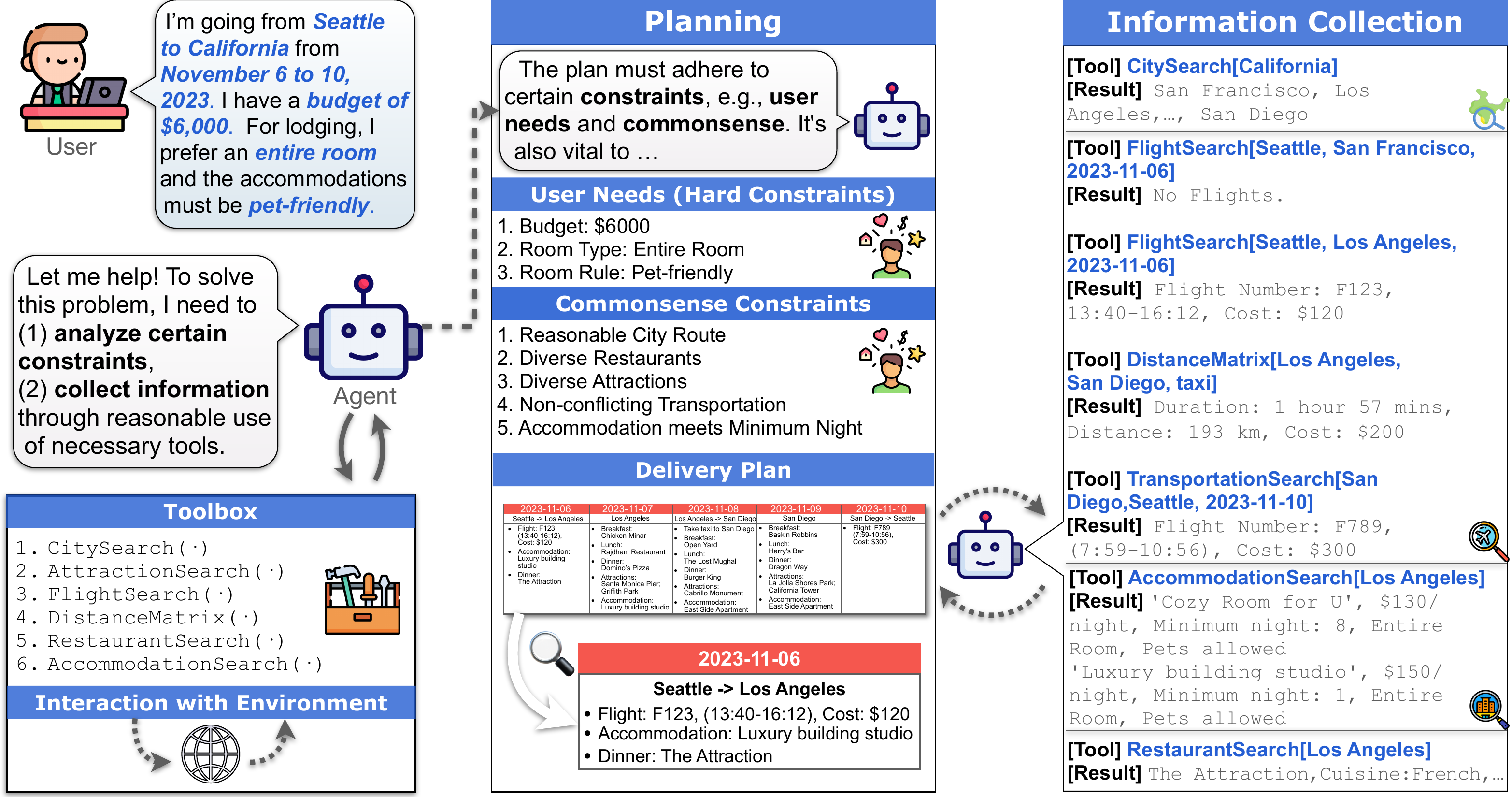}
    \vspace{-1.1em}
    \caption{Overview of \BenchmarkName. Given a query, language agents are tasked with employing various search tools to gather information. Based on the collected information, language agents are expected to deliver a plan that not only satisfies the user's needs specified in the query but also adheres to commonsense constraints.}
    \vspace{-1.2em}
    \label{fig:main}
\end{figure*}

Planning is a hallmark of human intelligence. 
It is an evolutionary feat built upon numerous other capacities: using various tools to iteratively collect information and make decisions, recording intermediate plans (in working memory or on a physical device) for deliberation, and exploring alternative plans by running simulations, which in turn depends on a world model~\cite{mattar2022planning, ho2022planning}.
For decades, researchers have been attempting to develop AI agents to mimic humans' planning capability~\cite{russell2010artificial,georgievski2015htn,karpas2020automated,lin2023decision,zhang2024timearena}, but often in constrained settings~\cite{campbell2002deep,silver2016mastering, silver2017predictron} because many of the cognitive substrates necessary for human-level planning have been lacking.
AI agents that can work robustly in the largely unconstrained settings in which humans operate remain a distant goal.

The advent of large language models (LLMs; \citet{openai2022chatgpt,openai2023gpt4,touvron2023llama,touvron2023llama2,jiang2023mistral}) brings new light to this classic problem. 
A new generation of language agents~\cite{su2023language,sumers2023cognitive,xie2023openagents} powered by LLMs has emerged, characterized by their capability of using language as a vehicle for thought and communication.
These agents have shown interesting capabilities, such as tool use~\cite{schick2023toolformer,patil2023gorilla,qin2023toolllm} and various forms of reasoning~\cite{wei2022chain,yao2022react,DBLP:conf/nips/LewkowyczADDMRS22}, potentially fulfilling the role of some of the cognitive substrates that were lacking in earlier AI agents. 
Researchers have therefore investigated their potential in an array of planning tasks ranging from classic planning settings like Blocksworld~\cite{Valmeekam2022PlanBenchAE} to embodied agents~\cite{huang2022language, DBLP:conf/corl/IchterBCFHHHIIJ22, song2023llm, wang2023voyager} and web agents~\cite{deng2023mindweb,zhou2023webarena}.
However, the planning settings in existing work still largely follow the conventional setting that focuses on single-objective optimization with fixed ground truths. 
An agent is tasked with predicting from a pre-defined set of actions, just now made by an LLM-powered agent.

\textit{Are language agents capable of planning in more complex yet realistic settings, closer to those in which humans operate?} To advance this investigation, we propose \BenchmarkName, a new planning benchmark that focuses on a common real-world planning scenario---travel planning.
This is a challenging, time-consuming task even for humans (but most people can do it successfully, with the right tools and enough time): \begin{inparaenum}[\it 1)]
\item Planning a multi-day itinerary is inherently \textit{long-horizon}, involving a large number of interdependent decisions on places, lodging, transportation, dining, etc.
\item Travel planning involves many \textit{constraints}, ranging from explicit constraints such as budget and various user needs to implicit commonsense constraints, e.g., people cannot teletransport to another city without using some means of transportation.
\item Travel planning requires strong \textit{agency} to proactively acquire necessary information using various tools (e.g., to search flights and restaurants) from the partially observable environment and deliberate over the collected information to further the planning while being mindful of all the explicit and implicit constraints.
\end{inparaenum}
Planning tasks of such complexity are out of the reach of prior AI agents~\cite{russell2010artificial}.

\BenchmarkName{} provides a rich sandbox environment with around four million data entries crawled from the Internet that can be accessed via six tools. 
We also meticulously curate \num{1225} diverse user queries (along with their reference plans), each imposing a different combination of constraints. 
A representative example is illustrated in Figure \ref{fig:main}. 

We comprehensively evaluate five LLMs, such as GPT-4~\cite{openai2023gpt4}, Gemini~\cite{team2023gemini}, and Mixtral~\cite{jiang2024mixtral}, and four planning strategies, such as ReAct~\cite{yao2022react} and Reflexion~\cite{shinn2023reflexion}, on their capability of delivering complete plans and following constraints. 
The main findings are as follows:

$\bullet$ State-of-the-art LLMs cannot handle complex planning tasks like those in \BenchmarkName. GPT-4 successfully produces a plan that meets all the constraints for a few tasks (\num{0.6}\%), while all other LLMs fail to complete any tasks.   

$\bullet$ Existing planning strategies such as ReAct and Reflexion, which may be effective for simpler planning settings, are insufficient for the multi-constraint tasks in \BenchmarkName. They often fail to convert their reasoning into the right actions correctly and keep track of global or multiple constraints. 
Language agents need more sophisticated planning strategies to approach human-level planning.

$\bullet$ Further analyses reveal many common failure modes of existing language agents, such as argument errors in tool use, being trapped in dead loops, and hallucinations.

Although most of our findings lean negatively toward the current language agents, we should note that the mere possibility for an artificial agent to tackle such a complex task is non-trivial progress in itself. 
\BenchmarkName{} provides a challenging yet meaningful testbed for future agents to hill-climb toward human-level planning in complex settings.

Finally, a silver lining: while our well-trained human annotators averagely take \num{12} minutes to manually annotate a plan, a language agent can produce a plan in just \num{1}--\num{2} minutes automatically. 
Perhaps one day, language agents will become capable enough to help automate away many of such tedious tasks for us.

%% file: 030relatedWork.tex
\subsection{Large Language Model based Agents}
\label{subsec:related-agent}
Empowered by large language models (LLMs), language agents have the capability to decompose complex tasks and arrive at solutions through a series of reasoned actions.
Notable examples such as AutoGPT~\cite{AutoGPT}, BabyAGI~\cite{BabyAGI}, and HuggingGPT~\cite{shen2023hugginggpt} have illuminated the community with their impressive abilities. 
Current LLM-powered language agents, equipped with Memory, Tool-use, and Planning modules, have seen a substantial improvement in their general abilities~\cite{weng2023prompt}.
Memory in language agents refers to their ability to acquire and process information. 
It is divided into two types: long-term memory, which is the parametric memory inherent in LLMs, and short-term memory, also known as in-context learning~\cite{brown2020language} or working memory.
Techniques like memory summarization ~\cite{chen2023walking,zhou2023recurrentgpt,liang2023unleashing} and retrieval~\cite{andreas-2022-language,park2023generative,zhong2023memorybank} are widely employed to enhance the memory capabilities of language agents.
Moreover, by interacting with external tools, language agents expand their potential capabilities significantly. 
This tool-augmentation paradigm has been validated as effective in previous work~\cite{nakano2021webgpt,lu2023chameleon,ge2023openagi,xie2023openagents}.
We further discuss the planning module in Section \ref{subsec:related-plan}.

\subsection{Planning}
\label{subsec:related-plan}
Planning, a hallmark of human intelligence, entails a sequence of actions that involve decomposing tasks, searching for solutions, and making final decisions~\cite{hayes1979cognitive,grafman2004planning,su2023language}.
This skill is crucial for achieving human-level intelligence and has been widely studied in areas such as robotics~\cite{mcdermott1992robot,alterovitz2016robot} and transportation scheduling~\cite{Cross1994DART,pinedo2005planning}.
The emergence of language agents powered by LLMs has further intensified discussions around their planning capabilities~\cite{liu2023llm+,Valmeekam2022PlanBenchAE}.
Previous research has demonstrated that language agents can effectively decompose tasks and engage in step-by-step reasoning, leading to significant improvements~\cite{wei2022chain,yuan-etal-2023-distilling,zheng2023seeact}.
Furthermore, to optimize solution searches in fewer steps, classical data structures like trees and graphs have been employed in prior studies~\cite{yao2023tree,besta2023graph}, enhancing the planning capabilities of language agents. 
In addition, methods involving feedback from the environment~\cite{yao2022react,shinn2023reflexion} have also been shown to be beneficial.
However, while these planning abilities have shown promise in specific tasks, the effectiveness of these planning strategies in scenarios with multiple constraints remains uncertain.

\subsection{Evaluation of Language Agents}
\label{subsec:related-eval}
Previous studies typically assess LLM-powered language agents in focused domains: arithmetic reasoning targeting correct solutions~\cite{roy-roth-2015-solving,cobbe2021training,patel-etal-2021-nlp}; tool-use evaluating agents' proficiency in employing tools and reporting results~\cite{li2023api,xu2023tool,zhuang2023toolqa}; and web navigation, testing agents' ability to locate specific websites~\cite{deng2023mindweb,zhou2023webarena,liu2023agentbench}.
However, the complexity of the real-world implies that previous evaluation methods, which focus on single objective and fixed ground truths, may fall short of capturing the full scope of agents' capabilities. 
To address this, we introduce \BenchmarkName~for comprehensive evaluations, assessing whether language agents can generate feasible solutions facing various objectives, referred to as constraints in this paper.

%% file: 040benchmark.tex
\subsection{Overview}
\label{subsec-benchmark-overview}
We introduce \BenchmarkName, a benchmark crafted for evaluating language agents in tool-use and complex planning within multiple constraints.
Grounding to travel planning, a real-world use-case that naturally includes diverse constraints such as user needs and commonsense constraints, \BenchmarkName~evaluates whether agents can develop flexible travel plans by collecting information via diverse tools and making decisions while satisfying the constraints.

\BenchmarkName~comprises \num{1225} queries in total.
The queries in \BenchmarkName~are divided into nine groups. 
This classification is based on two criteria: the duration of travel and the number of hard constraints.
The dataset is divided into the training, validation, and test set.
The training set includes \num{5} queries per group with human-annotated plans (\num{45} pairs in total), the validation set includes \num{20} queries per group (\num{180} in total), and the test set includes \num{1000} queries.
Detailed distributions are shown in Table \ref{tab:dataset-dist}.

\begin{table*}[t]
    \centering
    \vspace{-1em}
    \caption{Constraint description. The environment constraints are manifested through the feedback received from the environment, assessing whether the language agent can adjust its plan appropriately. The commonsense constraints and hard constraints are evaluated based on how well the language agent's plan aligns with these specific criteria.
    }
    \input{tables/constraint_description}
    \label{tab:cons-description}
    \vspace{-1em}
\end{table*}

\subsection{Constraint Introduction}
\label{subsec-benchmark-constraint}
In order to assess whether agents can perceive, understand, and satisfy various constraints to formulate a feasible plan, as outlined in Table \ref{tab:cons-description}, we include three types of constraints:

$\bullet$ \textbf{Environment Constraints}: The real-world is dynamic, necessitating agents to be adaptable.
For instance, flights to a particular destination may be unavailable at certain times (e.g., \textbf{{\color[HTML]{b66368} no flights from Seattle to San Francisco in Figure}} \ref{fig:main}), possibly because tickets are sold out.
In such cases, the agent must dynamically seek an alternative, like changing the destination of the flight or the way of transportation.
To simulate this, we introduce environment constraints within \BenchmarkName~to test the adaptability of agents in planning.

$\bullet$ \textbf{Commonsense Constraints}: Agents, functioning in real-world and serving humans, should consider commonsense when designing plans.
For instance, repeatedly visiting the same attraction is not typical.
To evaluate agents' understanding and utilization of commonsense during planning, we include the commonsense constraint in \BenchmarkName.

$\bullet$ \textbf{Hard Constraints}: A crucial ability for agents is to effectively satisfy personalized user needs.
To evaluate this, \BenchmarkName~incorporates various user needs, such as budget constraints.
These user needs are termed hard constraints.
The hard constraint measures the agent's generalization ability with regard to different user needs.

\begin{table}[t]
\vspace{-.8em}
    \centering
    \caption{The number of data entries in the database.}
    \input{tables/tool_database}
    \label{tab:tool-database}
    \vspace{-2em}
\end{table}

\subsection{Benchmark Construction Pipeline}
\label{subsec-benchmark-construction}
This section outlines the construction pipeline of \BenchmarkName, which involves the following steps:
\begin{inparaenum}[\it 1)]
\item Environment and evaluation setup.
\item Diverse travel query design.
\item Reference plan annotation.
\item Quality check.
\end{inparaenum}
\paragraph{Environment Setting.}
In \BenchmarkName, we create a static and closed sandbox environment for consistent and unbiased evaluations.
This setup ensures that all agents access the same unchanging information from our static databases, avoiding the variability and potential biases introduced by dynamic data. 
To offer various travel options that align with the real-world, we ensure the database for each tool in \BenchmarkName~contains rich information.
The database size of each tool is listed in Table \ref{tab:tool-database}.
For more tool details, please refer to Appendix \ref{appendix:tool-desc} and \ref{appendix:dataset-database}.
Additionally, agents are instructed to use the ``NotebookWrite'' tool to record necessary information for planning. 
This tool is integrated to evaluate agents' working memory management and prevents maximum token limit caused by context accumulation.
\paragraph{Query Construction.}
To create diverse queries for \BenchmarkName, we begin with several fundamental elements, including departure city, destination, and specific date range, randomly chosen to form the skeleton of each query.
Subsequently, we adjust the duration of travel and the number of hard constraints to create different levels of complexity.

The duration of the travel---\num{3}, \num{5}, or \num{7} days---determines the number of cities included in the plan.
Specifically, \num{3}-day plans focus on one city, while \num{5} days and \num{7} days involve visiting one randomly chosen state, with trips to \num{2} cities for the \num{5}-day plans and \num{3} cities for the \num{7}-day plans, respectively.
A greater number of days requires more frequent tool usage by language agents, thus increasing the difficulty of managing the long-horizon aspect of planning.
The uncertain destination challenges agents to decide on multiple cities, where they must consider factors such as inter-city connectivity.

Furthermore, we introduce diverse user needs as hard constraints to add further complexity and realism.
The difficulty levels are categorized as follows:

$\bullet$ \textbf{Easy}: Queries at this level are primarily budget-constrained for a single person. The initial budget for each query is determined using a set of crafted heuristic rules.

$\bullet$ \textbf{Medium}: In addition to budget constraints, medium queries introduce an additional hard constraint, randomly selected from a constraint pool including cuisine type, room type, and room rule.
Furthermore, the number of people varies between \num{2} and \num{8}, which influences the calculation of costs for both transportation and accommodation.

$\bullet$ \textbf{Hard}: Hard queries include additional transportation preference into the constraint pool, along with all the constraints in medium queries. 
Each hard query contains three hard constraints randomly selected from the constraint pool.

This method ensures the diversity of queries.
Minor changes in these elements can lead to significantly different plans.
Finally, based on elements, we utilize GPT-4~\cite{openai2023gpt4} to generate natural language queries.

\paragraph{Human Annotation.}
To ensure every query has at least one feasible plan, we invite \num{20} graduate students to meticulously annotate plans for synthesized queries.
One plan is deemed eligible only if it meets all the constraints outlined in our evaluation script, which is detailed in Section \ref{subsec-benchmark-evaluation}.
This rigorous process resulted in the creation of \num{1225} validated query-plan pairs.
We pay annotators an average of \$\num{0.80} for each plan they annotate.

\paragraph{Quality Control.}

To ensure the quality of each natural language query and its corresponding annotated plan, the authors performed a detailed review of every query and plan, rectifying any errors found.
Additionally, to ensure the challenges, we re-calibrate each query's budget using the costs from corresponding human-annotated plans. 
This approach replaces the initial heuristic-generated budgets, which might be too high, thus reducing the number of feasible plans.
Through multiple stages of human verification, we ensure the high quality of each query in \BenchmarkName~and the presence of at least one feasible solution.

\subsection{Evaluation}
\label{subsec-benchmark-evaluation}
To ensure a comprehensive evaluation of the plans offered by agents, we assess them from multiple dimensions. 
Specifically, we first extract key components\footnote{In our experiments, we use GPT-4-Turbo for this extraction process. Please refer to Appendix \ref{appendx:prompt-extraction} for more details.}, including transportation, restaurants, attractions, and accommodations, which are initially presented as natural language.
These components are then organized into a formally structured plan, which will be evaluated automatically through pre-defined scripts. 
The evaluation criteria include the following:

$\bullet$ \textbf{Delivery Rate}: This metric assesses whether agents can successfully deliver a final plan within a limited number of steps.
Falling into dead loops, experiencing numerous failed attempts, or reaching the maximum number of steps (\num{30} steps in our experimental setting) will result in failure.

$\bullet$ \textbf{Commonsense Constraint Pass Rate}: Comprising eight commonsense dimensions, this metric evaluates whether a language agent can incorporate commonsense into their plan without explicit instructions.

$\bullet$ \textbf{Hard Constraint Pass Rate}: This metric measures whether a plan satisfies all explicitly given hard constraints in the query, which aims to test the agents' ability to adapt their plans to diverse user needs.

\smallskip
$\bullet$ \textbf{Final Pass Rate}: This metric represents the proportion of feasible plans that meet all aforementioned constraints among all tested plans.
It serves as an indicator of agents' proficiency in producing plans that meet a practical standard.

We do not separately assess environment constraints since their impact is inherently reflected in the ``Within Sandbox'' and ``Complete Information'' metrics.
For instance, when cities lack transportation or attractions, agents typically resort to hallucination or opt not to provide an answer, reflecting the impact of environment constraints.

For the Commonsense Constraint Pass Rate and Hard Constraint Pass Rate, we utilize two evaluation strategies: \textit{micro} and \textit{macro}. 
The \textit{micro} strategy calculates the ratio of passed constraints to the total number of constraints.
The \textbf{Micro Pass Rate} is defined as:
\begin{equation}
    \text{Micro Pass Rate} = \frac{\sum_{p \in P}\sum_{c \in C_p} \mathbbm{1}_{\text{passed}(c, p)}}{\sum_{p \in P} |C_p|},
\end{equation}
where $P$ represents the set of all plans being evaluated, $C_{p}$ denotes the set of constraints applicable to a specific plan $p$ in $P$, and $\text{passed}(X,Y)$ is a function determining whether $Y$ meets constraints $X$.

The \textit{macro} strategy calculates the ratio of plans that pass all commonsense or hard constraints among all tested plans.
We define the \textbf{Macro Pass Rate} as:
\begin{equation}
    \text{Macro Pass Rate} = \frac{\sum_{p \in P} \mathbbm{1}_{\text{passed}(C_{p}, p)}}{|P|}.
\end{equation}
These two metrics evaluate an agent's capability of following individual constraints vs.\ all the constraints holistically.

\begin{table*}[t]
    \centering
    \vspace{-1em}
    \caption{Main results of different LLMs and planning strategies on the \BenchmarkName~validation and test set. The best results are marked in bold. When the collected information is insufficient, Gemini Pro tends to directly refuse to provide the plan. Interviews with annotators reveal that manually annotating a plan averagely takes around \num{12} minutes. However, language agents, such as GPT-3.5-Turbo, can accomplish this task in just \num{1} to \num{2} minutes, showcasing their efficiency.
    }
    \input{tables/main_res}
    \vspace{-1em}
    \label{tab:main-res}
\end{table*}

\subsection{Sole-Planning Setting}
While \BenchmarkName~is designed to assess the overall abilities of agents in tool-use and planning (two-stage mode), we also setup a simplified mode solely evaluating agents' planning skills (sole-planning mode).
In this setting, we utilize human-annotated plans to pre-determine the destination cities, and provide detailed and necessary information directly to agents, such as restaurants in the provided cities.
This eliminates the need for tool calling as agents don't need to collect information from scratch via tools anymore.

%% file: tables/constraint_description.tex
\resizebox{\linewidth}{!}{
\small
\begin{tabular}{ll}
\toprule
\textbf{Constraint}                         & \textbf{Description}                                                                                                                                                                           \\ \midrule
\rowcolor[gray]{0.85}
\multicolumn{2}{c}{\textbf{\textit{Environment Constraint}}}                                                                                                                                                                                 \\ \midrule
Unavailable Transportation                  & \parbox{0.85\linewidth}{There is no available flight or driving information between the two cities.}                                                                                                         \\
Unavailable Attractions                     & \parbox{0.85\linewidth}{There is no available attraction information in the queried city.}                                                                                                                                    \\ \midrule
\rowcolor[gray]{0.85}
\multicolumn{2}{c}{\textbf{\textit{Commonsense Constraint}}}                                                                                                                                                                                 \\ \midrule
Within Sandbox                     & \parbox{0.85\linewidth}{All information in the plan must be within the closed sandbox; otherwise, it will be considered a hallucination.}                                                                             \\
\addlinespace[0.25em]
Complete Information          & \parbox{0.85\linewidth}{No key information should be left out of the plan, such as the lack of accommodation during travel.}                                                                                       \\
\addlinespace[0.25em]
Within Current City                & \parbox{0.85\linewidth}{All scheduled activities for the day must be located within that day's city(s).}                                                                                                                         \\
\addlinespace[0.25em]
Reasonable City Route               & \parbox{0.85\linewidth}{Changes in cities during the trip must be reasonable.}                                                                                                                        \\
\addlinespace[0.25em]
Diverse Restaurants            & \parbox{0.85\linewidth}{Restaurant choices should not be repeated throughout the trip.}                                                                                                                  \\
\addlinespace[0.25em]
Diverse Attractions             & \parbox{0.85\linewidth}{Attraction choices should not be repeated throughout the trip.}                                                                                                                          \\
\addlinespace[0.25em]
Non-conf. Transportation      & \parbox{0.85\linewidth}{Transportation choices within the trip must be reasonable. For example, having both ``self-driving'' and ``flight'' would be considered a conflict.}\\
\addlinespace[0.5em]
Minimum Nights Stay & \parbox{0.85\linewidth}{The number of consecutive days spent in a specific accommodation during the trip must meet the corresponding required minimum number of nights' stay.}                                                                   \\ \midrule
\rowcolor[gray]{0.85}
\multicolumn{2}{c}{\textbf{\textit{Hard Constraint}}}                                                                                                                                                                                        \\ \midrule
Budget                             & \parbox{0.85\linewidth}{The total budget of the trip.} 
\\
\addlinespace[0.25em]
Room Rule                          & \parbox{0.85\linewidth}{Room rules include ``No parties'', ``No smoking'', ``No children under 10'', ``No pets'', and ``No visitors''.}                                                                                    \\
\addlinespace[0.25em]
Room Type                          & \parbox{0.85\linewidth}{Room types include ``Entire Room'', ``Private Room'', ``Shared Room'', and ``No Shared Room''.}                                                                                        \\
\addlinespace[0.25em]
Cuisine                            & \parbox{0.85\linewidth}{Cuisines include ``Chinese'', ``American'', ``Italian'', ``Mexican'', ``Indian'', ``Mediterranean'', and ``French''.}                                                                                 \\
\addlinespace[0.25em]
Transportation                     & \parbox{0.85\linewidth}{Transportation options include ``No flight'' and ``No self-driving''.}                                                                                                                              \\ \bottomrule
\end{tabular}}

%% file: tables/tool_database.tex
\small
\begin{tabular}{lll}
\toprule
\textbf{Tool}                 & \textbf{Data Entries} (\#)  \\ 
\midrule
CitySearch           & \num{312}        \\
FlightSearch         & \num{3827361}  \\
DistanceMatrix & \num{17603}     \\
RestaurantSearch     & \num{9552}      \\
AttractionSearch     & \num{5303}      \\
AccommodationSearch & \num{5064}      \\   
\bottomrule
\end{tabular}

%% file: tables/main_res.tex
\definecolor{CustomColor1}{RGB}{245,210,209}
\definecolor{CustomColor2}{RGB}{230,232,245} 
\definecolor{CustomColor3}{RGB}{235,235,237}

\newcolumntype{a}{>{\columncolor{CustomColor1}}c}
\newcolumntype{b}{>{\columncolor{CustomColor2}}c}
\newcolumntype{d}{>{\columncolor{CustomColor3}}c}
\setlength\tabcolsep{3pt} 
\resizebox{1.0\linewidth}{!}{
\begin{tabular}{lcccccccccccc}
\toprule

                             & \multicolumn{6}{c}{\textbf{Validation} (\#\num{180})}                                                                                                                                                                                                                                                                                                 & \multicolumn{6}{c}{\textbf{Test}  (\#\num{1000})}                                                                                                                                                                                                                                                                                                     \\ \cmidrule(l){2-7} \cmidrule(l){8-13}
                             & \multirow{2}{*}{\begin{tabular}[c]{@{}c@{}}Delivery \\ Rate\end{tabular}} & \multicolumn{2}{c}{\begin{tabular}[c]{@{}c@{}}Commonsense\\ Pass Rate\end{tabular}} & \multicolumn{2}{c}{\begin{tabular}[c]{@{}c@{}}Hard Constraint \\ Pass Rate\end{tabular}} & \multirow{2}{*}{\begin{tabular}[c]{@{}c@{}}Final \\ Pass Rate\end{tabular}} & \multirow{2}{*}{\begin{tabular}[c]{@{}c@{}}Delivery \\ Rate\end{tabular}} & \multicolumn{2}{c}{\begin{tabular}[c]{@{}c@{}}Commonsense \\ Pass Rate\end{tabular}} & \multicolumn{2}{c}{\begin{tabular}[c]{@{}c@{}}Hard Constraint \\ Pass Rate\end{tabular}} & \multirow{2}{*}{\begin{tabular}[c]{@{}c@{}}Final \\ Pass Rate\end{tabular}} \\ \cmidrule(l){3-4} \cmidrule(l){5-6}\cmidrule(l){9-10}\cmidrule(l){11-12}
                             &                                                                          & Micro                                    & Macro                                    & Micro                                       & Macro                                      &                                                                             &                                                                          & Micro                                     & Macro                                    & Micro                                       & Macro                                      &                                                                             \\ \midrule
\rowcolor{CustomColor3}
\multicolumn{1}{l}{\cellcolor{white}Greedy Search}                & \num{100}                                                                      & \num{74.4}                                     & \num{0}                                        & \num{60.8}                                        & \num{37.8}                                       & \num{0}                                                                           & \num{100}                                                                      & \num{72.0}                                      & \num{0}                                        & \num{52.4}                                        & \num{31.8}                                       & \num{0}                                                                           \\ \midrule
\rowcolor[gray]{0.85}
\multicolumn{13}{c}{\textbf{\textit{Two-stage}}}                                                                                                                                            \\ \midrule

\rowcolor{CustomColor2}
\multicolumn{1}{l}{\cellcolor{white}Mistral-7B-32K~\cite{jiang2023mistral}}             & \num{8.9}                                                     & \num{5.9}                     & \num{0}                     & \num{0}                        & \num{0}                       & \num{0}                                                        & \num{7.0}                                                     & \num{4.8}                      & \num{0}                     & \num{0}                        & \num{0}                       & \num{0}                                                        \\  
\rowcolor{CustomColor2}
\multicolumn{1}{l}{\cellcolor{white}Mixtral-8×7B-MoE~\cite{jiang2024mixtral}}             & \num{49.4}                                                     & \num{30.0}                     & \num{0}                     & \num{1.2}                        & \num{0.6}                & \num{0}       & \num{51.2}                                                        & \num{32.2}                                                     & \num{0.2}                      &  \num{0.7}                 & \num{0.4}                       & \num{0}                                                              \\
\rowcolor{CustomColor2}
\multicolumn{1}{l}{\cellcolor{white}Gemini Pro~\cite{team2023gemini}
}
& \num{28.9}                                                     & \num{18.9}                    & \num{0}                     & \num{0.5}                        & \num{0.6}                     & \num{0}                                                       & \num{39.1}                                                   & \num{24.9}                      & \num{0}                     & \num{0.6}                      & \num{0.1}                      & \num{0}                                                        \\
\rowcolor{CustomColor2}
\multicolumn{1}{l}{\cellcolor{white}GPT-3.5-Turbo~\cite{openai2022chatgpt}}               & \num{86.7}                                                                     & \num{54.0}                                     & \num{0}                                        & \num{0}                                         & \num{0}                                        & \num{0}                                                                           & \num{91.8}                                                     & \num{57.9}                      & \num{0}                     & \num{0.5}                        & \num{0.6}                       & \num{0}                                                        \\
\rowcolor{CustomColor2}
\multicolumn{1}{l}{\cellcolor{white}GPT-4-Turbo~\cite{openai2023gpt4}}
                  & \textbf{89.4}                                                     & \textbf{61.1}                     & \textbf{2.8}                     & \textbf{15.2}                        & \textbf{10.6}                       & \textbf{0.6}                                                        & \textbf{93.1}                                                    & \textbf{63.3}                      & \textbf{2.0}                    & \textbf{10.5}                        &  \textbf{5.5}                       & \textbf{0.6}                                                      \\
\midrule
\rowcolor[gray]{0.85}
\multicolumn{13}{c}{\textbf{\textit{Sole-planning}}}                                                                                                                                                                                                                                                                                                                                                                                                                                                                                                                                                                                                                                                            \\ \midrule
\rowcolor{CustomColor1}
\multicolumn{1}{l}{\cellcolor{white}Direct$_{\rm GPT-3.5-Turbo}$}
    & \num{100}                                                                      & \num{60.2}                                     & \num{4.4}                                      & \num{11.0}                                        & \num{2.8}                                        & \num{0}                                                                           & \num{100}                                                                      & \num{59.5}                                      & \num{2.7}                                      & \num{9.5}                                        & \num{4.4}                                        & \num{0.6}                                                                         \\
\rowcolor{CustomColor1}
\multicolumn{1}{l}{\cellcolor{white}CoT$_{\rm GPT-3.5-Turbo}$}
       & \num{100}                                                                      & \num{66.3}                                     & \num{3.3}                                      & \num{11.9}                                       & \num{5.0}                                        & \num{0}                                                                           & \num{100}                                                                      & \num{64.4}                                      & \num{2.3}                                      & \num{9.8}                                        & \num{3.8}                                        & \num{0.4}                                                                         \\
\rowcolor{CustomColor1}
\multicolumn{1}{l}{\cellcolor{white}ReAct$_{\rm GPT-3.5-Turbo}$}   & \num{82.2}                                                                     & \num{47.6}                                     & \num{3.9}                                      & \num{11.4}                                        & \num{6.7}                                        & \num{0.6}                                                                         &  \num{81.6}                                                                        &   \num{45.9}                                        &                \num{2.5}                          &      \num{10.7}                                       &                \num{3.1}                            &   \num{0.7}                                                                          \\
\rowcolor{CustomColor1}
\multicolumn{1}{l}{\cellcolor{white}Reflexion$_{\rm GPT-3.5-Turbo}$}
  & \num{93.9}                                                                     & \num{53.8}                                     & \num{2.8}                                      & \num{11.0}                                        & \num{2.8}                                        & \num{0}                                                                           & \num{92.1}                                                                         &     \num{52.1}                                      &      \num{2.2}                                    &     \num{9.9}                                        &      \num{3.8}                                      &    \num{0.6}                                                                         \\
  \rowcolor{CustomColor1}
\multicolumn{1}{l}{\cellcolor{white}Direct$_{\rm Mixtral-8{x}7B-MoE}$}             & \num{100}                                                                      & \num{68.1}                                     & \num{5.0}                                     &   \num{3.3}                                     & \num{1.1}                                       & \num{0}                                                                         & \num{99.3}                                                                      & \num{67.0}                                     &  \num{3.7}                                    & \num{3.9}                                        & \num{1.6}                                       & \num{0.7}                                                                         \\
\rowcolor{CustomColor1}
\multicolumn{1}{l}{\cellcolor{white}Direct$_{\rm Gemini~Pro}$}            & \num{93.9}                                                                      & \num{65.0}                                     & \num{8.3}                                     & \num{9.3}                                        & \num{4.4}                                       & \num{0.6}                                                                         & \num{93.7}                                                                     & \num{64.7}                                      & \num{7.9}                                     & \num{10.6}                                        & \num{4.7}                                       & \num{2.1}                                                                         \\

\rowcolor{CustomColor1}
\multicolumn{1}{l}{\cellcolor{white}Direct$_{\rm GPT-4-Turbo}$}            & \textbf{100}                                                                      & \textbf{80.4}                                     & \textbf{17.2}                                     & \textbf{47.1}                                        & \textbf{22.2}                                       & \textbf{4.4}                                                                         & \textbf{100}                                                                      & \textbf{80.6}                                      & \textbf{15.2}                                     & \textbf{44.3}                                        & \textbf{23.1}                                       & \textbf{4.4}                                                                         \\
\bottomrule
\end{tabular}}

%% file: 050experiments.tex
We evaluate the performance of various LLMs and planning strategies on \BenchmarkName. 
In the two-stage mode, we use the ReAct~\cite{yao2022react} framework for information collection, which is recognized for its effective iteration with tools~\cite{zhuang2023toolqa} while varying the foundation LLMs.
This approach allows us to assess how different LLMs perform under a uniform tool-use framework.
The agents are required to give the plan directly based on the information collected by themselves, without employing any other planning strategies.
In the sole-planning mode, our evaluation goes beyond varying LLMs to include different planning strategies.
This aims to assess if the strategies proven effective in other planning benchmarks maintain their efficacy in \BenchmarkName.
All experiments are conducted in a zero-shot setting.

\subsection{Baselines}
\paragraph{Greedy Search.}
To evaluate the effectiveness of traditional rule-based strategies within \BenchmarkName, we include greedy search as a baseline and  set cost as the optimization objective. 
Please refer to Appendix \ref{appendix:experiment-baseline} for more details.

\paragraph{LLMs.} 
Due to the long context window requirement of ReAct and the massive information as text, we limit our consideration to LLMs capable of handling inputs exceeding \num{8}K in length.
As a result, our selection includes three closed-source LLMs: \textbf{GPT-3.5-Turbo}~\cite{openai2022chatgpt}, \textbf{GPT-4-Turbo}~\cite{openai2023gpt4}, and \textbf{Gemini Pro}~\cite{team2023gemini}, as well as two open-source LLMs: \textbf{Mistral-7B-32K}~\cite{jiang2023mistral} and \textbf{Mixtral-8x7B-MoE}~\cite{jiang2024mixtral}.
For all these models, we adopt the official instruction formats whenever available.

\paragraph{Planning Strategies.} 
To explore the effectiveness of current planning strategies, we evaluate four representative ones: 
\textbf{Direct}, \textbf{ZS-CoT}~\cite{wei2022chain}, \textbf{ReAct}~\cite{yao2022react}, and \textbf{Reflexion}~\cite{shinn2023reflexion}.
For the implementation details, please refer to Appendix \ref{appendix:experiment-baseline}.
We do not include \textbf{ToT}~\cite{yao2023tree} and \textbf{GoT}~\cite{besta2023graph} because they require extensive exploration of the search space, prohibitively costly for problems as complex as \BenchmarkName. 
Also, given their performance close to ReAct in complex tasks~\cite{zhuang2023toolchain}, the potential benefits of these methods may be limited.

\subsection{Main Results}
In this section, we discuss the performance of various LLMs and planning strategies on \BenchmarkName~(Table \ref{tab:main-res}).
We have the following observations:

\textbf{\BenchmarkName~poses a significant challenge.}
In the two-stage mode, GPT-4-Turbo with ReAct achieves only \num{0.6}\% in the final pass rate, and none of the other LLMs can pass any of the tasks.
Even given all the necessary information in the sole-planning mode, 
existing planning strategies like ReAct and Reflexion still struggle with planning in \BenchmarkName{,} even though they have shown their effectiveness in more conventional planning tasks.
It is noteworthy that the best-performing agent still falls short on hard constraints even when compared to greedy search. 
This poor performance underlines the difficulty of \BenchmarkName~and shows that current agents still struggle with complex planning. 

\textbf{Agents show a substantial gap between the two modes.} 
The comparison of the two modes reveals the agents' struggles in fiddling with both information collection and planning.
Across all metrics, the scores of any model in the two-stage mode are lower than those in the sole-planning mode, with the largest gap reaching over \num{30}\%.
Similar to humans, language agents also seem to have a limited ``cognitive capacity'' and their performance deteriorates when multitasking.  
We provide a further analysis in Section \ref{sec:planning-error}.

\textbf{Agents struggle in obtaining a high macro pass rate.}
While some agents achieve high micro scores, their macro scores remain low.
This pattern shows that although agents manage to satisfy some constraints, they often overlook some other constraints in the meantime. 
Consequently, this indicates the current agents fail to consider multiple constraints holistically, a critical requirement for navigating the complex tasks in \BenchmarkName.

In summary, \BenchmarkName~poses a great challenge to current agents. 
The SoTA LLMs and planning strategies, which often show equal or superior to human-level performance on many traditional tasks, are still far from sufficient for complex planning tasks that humans are capable of.
\BenchmarkName~provides a challenging yet meaningful benchmark for the development of more capable language agents.

%% file: 060analysis.tex
\subsection{Tool-Use Error Analysis}
\begin{figure}[t]
\centering
    \includegraphics[width=\linewidth]{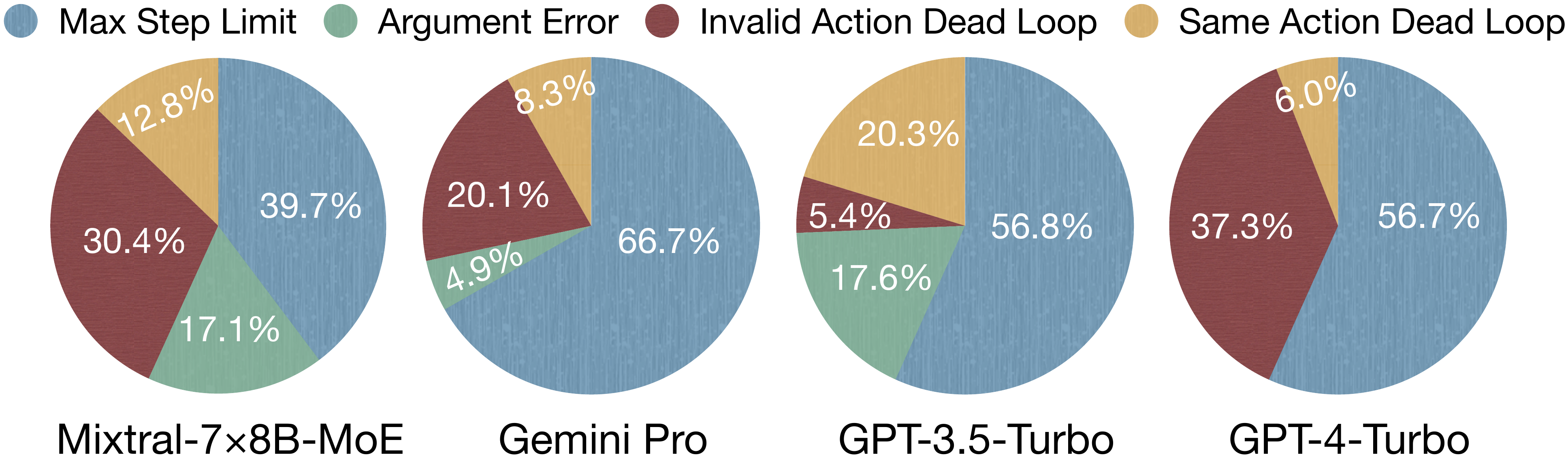}
    \vspace{-1.5em}
    \caption{Tool-use error distribution on the test set. An early stop will be triggered if the agent either makes three consecutive failed attempts or repetitive actions, indicating a dead loop.}
    \vspace{-1em}
    \label{fig:tool-use-error}
\end{figure}
As shown in Table \ref{tab:main-res}, even based on GPT-4-Turbo, agents still make mistakes in the process of information collection and thus fail to deliver a plan.
This problem is more severe in Gemini Pro and Mixtral.
To delve into the underlying causes, we categorize all error types in Figure \ref{fig:tool-use-error}.
We find:
\begin{inparaenum}[\it 1)]
\item \textbf{Agents incorrectly use tools.}
Except for GPT-4-Turbo, other LLMs-based agents all have argument error problems to varying degrees.
It sheds light that the use of simple tools still poses a significant challenge for agents.
\item \textbf{Agents trap in dead loops.}
Even with GPT-4-Turbo, invalid actions and repetitive action loops contribute to \num{37.3}\% and \num{6.0}\% of errors, respectively. 
Despite receiving feedback that actions are invalid or yield null results, agents persistently repeat these actions.
This suggests that agents fail to dynamically adjust their plans based on environment feedback.
\end{inparaenum}

\begin{table}[t]
    \centering
    \vspace{-1em}
    \caption{Constraint pass rate of GPT-4-Turbo on test set. The results of the sole-planning mode are based on the Direct strategy. 
}
    \input{tables/constraint_pass_rate}

    \label{tab:cons-pass-res}
    \vspace{-1em}
\end{table}
\begin{table}[t]
    \centering
    \vspace{-.5em}
    \caption{Comparison of the numbers of different tool uses between agent (GPT-4-Turbo) and reference. The results of agent are based on the number of entries written into the ``Notebook''.}
    \input{tables/agent_vs_human}
    \vspace{-1em}
    \label{tab:agent-vs-human}
\end{table}

\begin{figure*}[t]
    \centering
    \includegraphics[width=\linewidth]{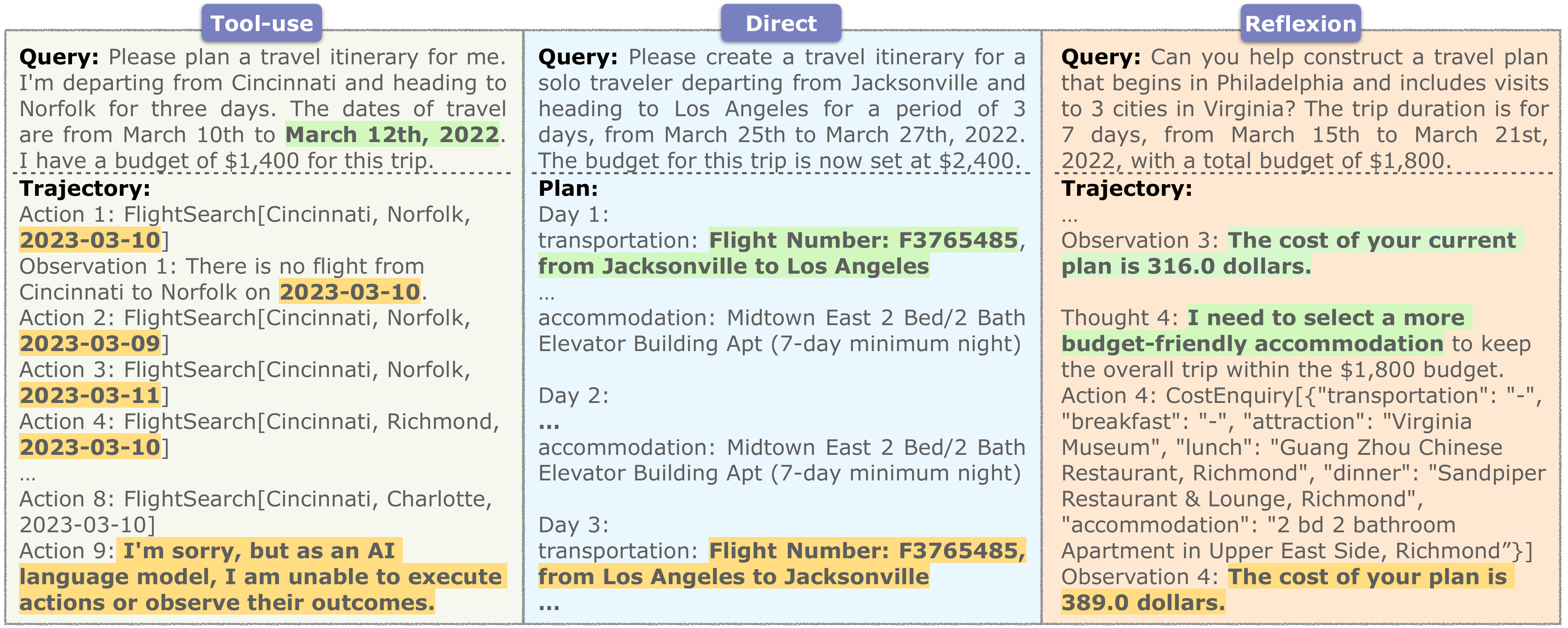}
    \vspace{-2.5em}
    \caption{Case studies of failures. Agents fail to finalize plans due to repeated errors like incorrect dates, confusion with information details leading to hallucinatory answers, and disconnects between reasoning and actions. All cases are gained from GPT-4-Turbo based agents. 
    For details of GPT-4-Turbo with the Reflexion strategy, please refer to Appendix \ref{appendix:experiment-gtp-4-reflexion}. We provide more cases in Appendix \ref{appendix:case-study}. 
    }
    \vspace{-1em}
    \label{fig:case-study}
\end{figure*}

\subsection{Planning Error Analysis}
\label{sec:planning-error}
We detail each constraint pass rate in Table \ref{tab:cons-pass-res}, from which we have the following observations:

\textbf{The number of hard constraints affects the performance of agents.}
Agents consistently exhibit pass rates below \num{10}\% across all levels of difficulty, and this performance deteriorates further as more constraints are introduced.
This trend highlights that current agents struggle with multi-constraint tasks, a key aspect of \BenchmarkName.

\textbf{Comprehensive information collection is essential for agents to plan effectively.}
In the sole-planning mode, agent performance improves over the two-stage mode.
To investigate this, Table \ref{tab:agent-vs-human} shows agents in two-stage mode use tools less effectively compared to the reference plans.
This comparison indicates that agents often fail to finish completed information collection.
Consequently, they either generate made-up information or omit specific details.
This leads to low pass rates for the ``Within Sandbox'' and ``Complete Information'' constraints. 
Additionally, this discrepancy becomes more pronounced with an increase in the duration of travel.
This emphasizes the need for agents to improve their capabilities in long-horizon tasks.

\textbf{Agents struggle with global planning scenarios.}
Global constraints ``Minimum Nights Stay'' and ``Budget'' demand a holistic approach to planning, necessitating that agents not only assess their current decisions but also anticipate future implications. 
Current LLMs' auto-regressive nature limits them to independently obtain outcomes from multiple future branches.
This highlights the necessity and urgent need of new strategies, such as backtracking for adjusting or employing heuristic methods for forward-looking planning.

\subsection{Case Studies}

To investigate the drawbacks of current agents in-depth, we provide several failure cases in Figure \ref{fig:case-study}.
We conclude with the following features:

\textbf{Agents fail to complete a plan due to the inability to rectify persistent errors.} 
In tool-use scenarios, agents often fail to deliver a plan even when all preceding steps are executed correctly. 
Further investigation reveals that this issue often stems from incorrect date inputs. 
As shown in the left part of Figure \ref{fig:case-study}, despite correct execution, agents repeatedly use incorrect dates.
This leads to null results, as the data in the \BenchmarkName~sandbox is based on 2022.
Such repeated failures eventually cause the agents to stop planning.
This indicates a significant limitation: current agents cannot self-correct their initial and incorrect assumptions.

\textbf{Agents produce hallucinatory answers due to information confusion.}
To understand why agents provide hallucinatory answers even when supplied with sufficient information in the sole-planning mode, we conduct a detailed analysis.
We observe a tendency for agents to confuse one piece of information with another. 
As shown in the middle part of Figure \ref{fig:case-study}, agents  mistakenly use the same flight number for both departure and return flights.
Such errors result in hallucinations, as the information provided in the plan does not align with the data in the sandbox.
This suggests that agents might be lost when faced with mass information, known as ``Lost in the Middle''~\cite{liu-etal:2023:arxiv}.

\textbf{Agents struggle to align their actions with their reasoning.}
To understand the reasons behind the lower delivery rate of Reflexion~\cite{shinn2023reflexion}, we examine specific examples. 
As illustrated in the right part of Figure \ref{fig:case-study}, we observe a discrepancy between what agents think and what they do. 
Despite recognizing the necessity to minimize costs, they tend to randomly select items, some of which may be more expensive. 
This discrepancy demonstrates that agents struggle to synchronize their actions with their analytical reasoning, severely impeding their delivery rate.

%% file: tables/constraint_pass_rate.tex
\definecolor{CustomColor1}{RGB}{245,210,209}
\definecolor{CustomColor2}{RGB}{230,232,245} 
\definecolor{CustomColor3}{RGB}{235,235,237} 

\setlength\tabcolsep{2pt} 

\newcolumntype{a}{>{\columncolor{CustomColor1}}c}
\newcolumntype{b}{>{\columncolor{CustomColor2}}c}
\newcolumntype{d}{>{\columncolor{CustomColor3}}c}
\resizebox{\linewidth}{!}{
\small
\begin{tabular}{laaabbb}
\toprule
\multirow{2}{*}{\textbf{Constraint Type}}   & \multicolumn{3}{c}{\textbf{Two-stage}} & \multicolumn{3}{c}{\textbf{Sole-planning}} \\ \cmidrule(l){2-4} \cmidrule(l){5-7}
                                   & \multicolumn{1}{c}{Easy}     & \multicolumn{1}{c}{Medium}    & \multicolumn{1}{c}{Hard}    & \multicolumn{1}{c}{Easy}      & \multicolumn{1}{c}{Medium}     & \multicolumn{1}{c}{Hard}     \\ \midrule
\rowcolor[gray]{0.85}
\multicolumn{7}{c}{\textbf{\textit{Commonsense Constraint}}}                                                              \\ \midrule
Within Sandbox & \num{37.4} & \num{31.2} & \num{33.9} & \num{76.4} & \num{71.5} & \num{79.3} \\
Complete Information & \num{53.4} & \num{52.9} & \num{58.0} & \num{94.5} & \num{96.4} & \num{96.2} \\
Within Current City & \num{69.3} & \num{67.3} & \num{68.3} & \num{89.1} & \num{80.8} & \num{82.4} \\
Reasonable City Route & \num{44.5} & \num{45.6} & \num{54.9} & \num{99.4} & \num{99.7} & \num{99.1} \\
Diverse Restaurants & \num{85.1} & \num{81.4} & \num{86.8} & \num{91.1} & \num{89.8} & \num{87.8} \\
Diverse Attractions & \num{94.3} & \num{90.4} & \num{94.0} & \num{100.0} & \num{100.0} & \num{100.0} \\
Non-conf. Transportation & \num{70.1} & \num{73.3} & \num{83.1} & \num{60.1} & \num{56.5} & \num{87.5} \\
Minimum Nights Stay & \num{46.8} & \num{46.2} & \num{51.1} & \num{37.4} & \num{28.8} & \num{30.1} \\
\midrule
\rowcolor[gray]{0.85}
\multicolumn{7}{c}{\textbf{\textit{Hard Constraint}}}                                                                     \\ \midrule
Budget                          & \num{10.1}      & \num{8.4}       & \num{4.4}     & \num{37.4}      & \num{35.1}       & \num{25.1}     \\
Room Rule                          & -        & \num{5.6}       & \num{11.3}     & -         & \num{31.5}       & \num{43.6}     \\
Cuisine                            & -        & \num{10.8}      & \num{11.4}    & -         & \num{57.5}       & \num{46.7}     \\
Room Type                          & -        & \num{12.4}       & \num{13.8}     & -         & \num{45.7}       & \num{56.7}     \\
Transportation                      & -        & -         & \num{18.6}    & -         & -          & \num{77.5}     \\ 
\midrule
\rowcolor[gray]{0.85}
\multicolumn{7}{c}{\textbf{\textit{Final}}} \\ 
\midrule
Final Pass Rate & \num{1.1}        & \num{0.3}         & \num{0.3}    & \num{8.0}         & \num{2.7}          & \num{2.2} 
\\ \bottomrule
\end{tabular}}

%% file: tables/agent_vs_human.tex
\setlength\tabcolsep{1.5pt} 
\resizebox{.85\linewidth}{!}{
\begin{tabular}{lcccccc}
\toprule
\multirow{2}{*}{\textbf{Average}} & \multicolumn{3}{c}{\textbf{Agent}} & \multicolumn{3}{c}{\textbf{Reference}} \\ \cmidrule(l){2-4} \cmidrule(l){5-7}
                      & 3-day   & 5-day  & 7-day  & 3-day   & 5-day  & 7-day  \\ \midrule
FlightSearch & 0.7    & \num{1.2}    & \num{0.8}    & \num{2.0}       & \num{3.0}      & \num{4.0}     \\
DistanceMatrix & \num{0.3}    & \num{0.6}    & \num{1.2}    & \num{2.0}       & \num{3.0}      & \num{4.0}     \\
RestaurantSearch     & \num{0.9}     & \num{1.5}    & \num{2.4}    & \num{1.0}       & \num{2.0}      & \num{3.0}      \\
AttractionSearch     & \num{0.9}     & \num{1.7}   & \num{2.4}    & \num{1.0}       & \num{2.0}      & \num{3.0}      \\
AccommodationSearch  & \num{0.9}    & \num{1.6}   & \num{2.5}    & \num{1.0}       & \num{2.0}      & \num{3.0}      \\ 
\bottomrule
\end{tabular}}

%% file: 070conclusion.tex
We introduce \BenchmarkName, a benchmark grounded in real-world scenarios, designed to assess the multi-constraint planning and tool-use abilities of current language agents.
Our benchmark presents a significant challenge: even the most advanced language agent frameworks only achieve a mere 0.6\% score in the final pass rate.
Further analysis shows that these agents are unable to take all constraints into consideration to deliver feasible plans.

\BenchmarkName's intricate logic and general applicability stand as vital components in the progressive development of language agents, thus contributing to the broader quest for AI abilities.
We envision \BenchmarkName~as a catalyst for future research, aiming to enhance agents' performance in increasingly complex scenarios, hill-climbing towards human-level cognitive capabilities.

%% file: 080impact.tex
\BenchmarkName~aims to provide an effective benchmark for complex planning in future research. 
Some of the data in the \BenchmarkName{} environment is derived from publicly available data on the Internet, and the content involved does not represent  the authors' viewpoints.
We realize that everyone's definition of commonsense may be different.
Our current evaluation criteria are based on the authors' consensus, and we encourage additional discussions to enrich our commonsense dimension, aiming for a more thorough evaluation.
We will release our evaluation scripts to foster innovation and aid the development of new methods.
We encourage the use of evaluation feedback in training set, such as implementing reinforcement learning techniques, to enhance learning.
However, we strictly prohibit any form of cheating in the validation and test sets to uphold the fairness and reliability of the benchmark's evaluation process.

%% file: 090appendix.tex
\setcounter{table}{0}
\renewcommand\thetable{\Alph{section}.\arabic{table}}
\setcounter{figure}{0}
\renewcommand\thefigure{\Alph{section}.\arabic{figure}}

Within this supplementary material, we elaborate on the following aspects:
\begin{compactitem}
\vspace{0.5em}
\item Appendix \ref{appendix:dataset}: Benchmark Details
\vspace{0.5em}
\item Appendix \ref{appendix:experiment}: Experiment Details
\vspace{0.5em}
\item Appendix \ref{appendix:case}: Case Presentation
\vspace{0.3em}
\end{compactitem}

\section{Benchmark Details}
\label{appendix:dataset}

\subsection{Dataset Distribution}
\label{appendix:dataset-dist}
In Table \ref{tab:dataset-dist}, we list the detailed group distribution on training, validation and test set.
\begin{table}[h]
    \centering
    \caption{Dataset distribution.}
    \input{tables/dataset_distribution}
    \label{tab:dataset-dist}
\end{table}

\subsection{Tool Description}
\label{appendix:tool-desc}
In Table \ref{tab:tool-dsecription}, we list the detailed tool description.  The original data for each tool is sourced from publicly available Internet data.  We then modify this data, which includes adding, deleting, and altering certain keys and values to suit our requirements. In this way, we effectively avoid the problem of data contamination. For more details, please refer to Appendix \ref{appendix:dataset-database}.
\begin{table*}[h]
    \centering
    \caption{Tool description and the number of data entries in the database.}
    \input{tables/tool_description}
    \label{tab:tool-dsecription}
    \vspace{-1em}
\end{table*}

\subsection{Environment Database Construction}
\label{appendix:dataset-database}
\paragraph{FlightSearch}
For FlightSearch, we source original data from the Kaggle Flight Status Prediction dataset\footnote{\url{https://www.kaggle.com/datasets/robikscube/flight-delay-dataset-20182022?select=Combined_Flights_2022.csv}}. 
From this dataset, we extract data spanning from 2022-03-01 to 2022-04-01. 
We specifically included fields like ``FlightDate'', ``DepTime'', ``ArrTime'', ``ActualElapsedTime'', ``Distance'', ``OriginCityName'', and ``DestCityName'' while discarding other values. 
To incorporate ``Price'' into our dataset, we generate this value by multiplying the ``Distance'' by a random factor ranging from 0.2 to 0.5.

\paragraph{DistanceMatrix}
We utilize the Google Distance Matrix API\footnote{\url{https://developers.google.com/maps/documentation/distance-matrix/overview?hl=en}} to calculate the driving distance and estimated travel time between two cities. 
For the ``self-driving'' and ``taxi'' modes of transportation, we calculate the 'Price' by multiplying the 'Distance' by factors of 1 and 0.15, respectively. 
To ensure consistency and reliability of data, we store the search results in our database, thereby creating a fixed dataset for our evaluations.

\paragraph{RestaurantSearch}
Our restaurant data is sourced from the Kaggle Zomato Restaurants Dataset\footnote{\url{https://www.kaggle.com/datasets/shrutimehta/zomato-restaurants-data}}. 
From this dataset, we extract the ``Restaurant Name'' and ``Average Cost'' for each establishment. 
Subsequently, we randomly assign these restaurants to various cities. 
To align with the constraint requirements of \BenchmarkName, we also randomly categorize each restaurant under the following cuisines: ``Chinese'', ``American'', ``Italian'', ``Mexican'', ``Indian'',``Mediterranean'', ``Middle Eastern'', ``Korean'', ``Asian'', ``French''.  

\paragraph{AttractionSearch}
For AttractionSearch, we employ the Google Places API\footnote{\url{https://developers.google.com/maps/documentation/places/web-service/overview?hl=en}} to gather information about attractions in each city. 
In \BenchmarkName, we retain essential details such as ``Name'', ``Address'', ``Phone'', ``Website'', ``Latitude'', and ``Longtitue'' for each attraction. 
To maintain data consistency and reliability, we store these search results in our database, creating a standardized dataset for our analyses.

\paragraph{AccommodationSearch}
Our accommodation data is obtained from the Kaggle Airbnb Open Data Dataset\footnote{\url{https://www.kaggle.com/datasets/arianazmoudeh/airbnbopendata}}. 
From this dataset, we extract key details ``NAME'',	``room type'',	``price'',	``minimum nights'',	``review rate number'',	and ``maximum occupancy''. 
Items are then randomly assigned to various cities. 
To meet the specific constraint requirements of \BenchmarkName, we also assign each item random room rules, including ``No parties'', ``No smoking'', ``No children under 10'', ``No pets'', and ``No visitors''.

\section{Experiment Details}
\label{appendix:experiment}

\subsection{Baselines}
\label{appendix:experiment-baseline}
\paragraph{Greedy Search}
To assess the effectiveness of traditional search algorithms in \BenchmarkName, we integrate a greedy search approach, focusing on minimizing costs. 
For 5 or 7-day travel plans, the first one or two cities in the returned city search result are selected as destinations.
The transportation choice is based on the lowest cost option among flights, taxis, and self-driving. 
The diet component involves selecting the restaurant with the lowest average cost. 
The cheapest accommodation is chosen for lodging. 
For attractions, we opt for a random selection for each day of the itinerary. 

\paragraph{Planning Strategy}
Current planning strategies have shown effectiveness in traditional tasks like mathematical problem-solving, but their capability to handle the more complex and constrained scenarios like \BenchmarkName~remains to be seen. 
To explore this, we evaluate four distinct planning strategies on \BenchmarkName:
\begin{inparaenum}[\it 1)]
\item \textbf{Direct}: In this method, the query is input directly into the model along with instructions detailing the task and relevant information gathered.
\item \textbf{ZS-CoT}~\cite{wei2022chain}: This strategy enhances the reasoning process by requiring intermediate steps.
Building on the Direct method, we add the prompt ``Let's think step by step'' to elicit reasoning.
\item \textbf{ReAct}~\cite{yao2022react}: This strategy incorporates environmental feedback into the reasoning process.
Thus, it enhances the language agent's reasoning ability by offering additional information.
In \BenchmarkName, we provide the cost associated with each entire day's plan as environmental feedback.
\item \textbf{Reflexion}~\cite{shinn2023reflexion}: This approach utilizes a reflection model to provide high-level insights on previous erroneous attempts.
Such reflective guidance aids language agents in identifying and correcting flawed reasoning.
\end{inparaenum}
In order to control the cost, we conduct tests on Direct using four different models, while the other strategies are evaluated using GPT-3.5-Turbo.
Detailed instructions for each strategy are available in Appendix \ref{appendix:experiment-prompt}.

\subsection{GPT-4-Turbo with Reflexion strategy in sole-planning mode.}
\label{appendix:experiment-gtp-4-reflexion}
We provide the results of GPT-4-Turbo with Reflexion strategy on validation set in Table \ref{tab:gpt4-relexion}.
\begin{table}[h]
    \centering
    \caption{GPT-4-Turbo with Reflexion strategy on validation set.}
    \input{tables/gpt4_relfexion}
    \label{tab:gpt4-relexion}
\end{table}

\subsection{Prompt List}
\label{appendix:experiment-prompt}

\subsubsection{Tool-use Prompt}
We tailor the ReAct~\cite{yao2022react} framework to suit our specific requirements in \BenchmarkName. 
An example of the instruction prompt for our needs is as follows:

\lstset{
    backgroundcolor=\color[RGB]{245,245,245},
    breaklines=true,
    breakindent=0pt,
    basicstyle=\ttfamily\small,
    frame=trbl,
    frameround = tttt,
}\begin{lstlisting}
Collect information for a query plan using interleaving 'Thought', 'Action', and  'Observation' steps. Ensure you gather valid information related to transportation, dining, attractions, and  accommodation. All information should be written in Notebook, which will then be input into the Planner  tool. Note that the nested use of tools is prohibited. 'Thought' can reason about the current situation, and 'Action' can have 8 different types:
(1) FlightSearch[Departure City, Destination City, Date]:
Description: 
A flight information retrieval tool.
Parameters: 
Departure City: The city you'll be flying out from.
Destination City: The city you aim to reach.
Date: The date of your travel in YYYY-MM-DD format.
Example: FlightSearch[New York, London, 2022-10-01] would fetch flights from New York to 
London on October 1, 2022.

(2) DistanceMatrix[Origin, Destination, Mode]:
Description: Estimate the distance, time and cost between two cities.
Parameters:
Origin: The departure city of your journey.
Destination: The destination city of your journey.
Mode: The method of transportation. Choices include 'self-driving' and 'taxi'.
Example: DistanceMatrix[Paris, Lyon, self-driving] would provide driving distance, time 
and cost between Paris and Lyon.

(3) AccommodationSearch[City]:
Description: Discover accommodations in your desired city.
Parameter: City - The name of the city where you're seeking accommodation.
Example: AccommodationSearch[Rome] would present a list of hotel rooms in Rome.

(4) RestaurantSearch[City]:
Description: Explore dining options in a city of your choice.
Parameter: City - The name of the city where you're seeking restaurant.
Example: RestaurantSearch[Tokyo] would show a curated list of restaurants in Tokyo.

(5) AttractionSearch[City]:
Description: Find attractions in a city of your choice.
Parameter: City - The name of the city where you're seeking attractions.
Example: AttractionSearch[London] would return attractions in London.

(6) CitySearch[State]
Description: Find cities in a state of your choice.
Parameter: State - The name of the city where you're seeking cities.
Example: CitySearch[California] would return cities in California.

(7) NotebookWrite[Short Description]
Description: Writes a new data entry into the Notebook tool with a short description. This tool should be used immediately after FlightSearch, AccommodationSearch, AttractionSearch, RestaurantSearch or DistanceMatrix. Only the data stored in Notebook can be seen by Planner. So you should write all the information you need into Notebook.
Parameters: Short Description - A brief description or label for the stored data. 
You don't need to write all the information in the description. 
The data you've searched for will be automatically stored in the Notebook.
Example: NotebookWrite[Flights from Rome to Paris in 2022-02-01] would store the 
informatrion of flights from Rome to Paris in 2022-02-01 in the Notebook.

(8) Planner[Query]
Description: A smart planning tool that crafts detailed plans based on user input and the information stroed in Notebook.
Parameters: 
Query: The query from user.
Example: Planner[Give me a 3-day trip plan from Seattle to New York] would return a 
detailed 3-day trip plan.
You should use as many as possible steps to collect engough information to input to the 
Planner tool. 

Each action only calls one function once. Do not add any description in the action.

Query: {query}
\end{lstlisting}

\subsubsection{Direct Planning Prompt}
We provide the instruction prompt of Direct strategy as follows:

\lstset{
    backgroundcolor=\color[RGB]{245,245,245},
    breaklines=true,
    basicstyle=\ttfamily\small,
    frame=trbl,
    frameround = tttt,
}\begin{lstlisting}
You are a proficient planner. Based on the provided information and query, please give me a detailed plan, including specifics such as flight numbers (e.g., F0123456), restaurant names, and accommodation names. Note that all the information in your plan should be derived from the provided data. You must adhere to the format given in the example. Additionally, all details should align with commonsense. The symbol '-' indicates that information is unnecessary. For example, in the provided sample, you do not need to plan after returning to the departure city. When you travel to two cities in one day, you should note it in the 'Current City' section as in the example (i.e., from A to B). 

***** Example *****
Query: Could you create a challenging travel plan for 7 people from Roanoke to Illinois spanning a week, from March 8th to March 14th, 2022, with a budget of $30,200? The preference is for an entire room, and we would not be taking any flights. In terms of cuisine, we are interested in sampling some Italian and Chinese food.
Travel Plan:
Day 1:
Current City: from Ithaca to Charlotte
Transportation: Flight Number: F3633413, from Ithaca to Charlotte, Departure Time: 05:38, 
Arrival Time: 07:46
Breakfast: Nagaland's Kitchen, Charlotte
Attraction: The Charlotte Museum of History, Charlotte
Lunch: Cafe Maple Street, Charlotte
Dinner: Bombay Vada Pav, Charlotte
Accommodation: Affordable Spacious Refurbished Room in Bushwick!, Charlotte

Day 2:
Current City: Charlotte
Transportation: -
Breakfast: Olive Tree Cafe, Charlotte
Attraction: The Mint Museum, Charlotte;Romare Bearden Park, Charlotte.
Lunch: Birbal Ji Dhaba, Charlotte
Dinner: Pind Balluchi, Charlotte
Accommodation: Affordable Spacious Refurbished Room in Bushwick!, Charlotte

Day 3:
Current City: Charlotte
Transportation: Flight Number: F3786167, from Charlotte to Ithaca, 
Departure Time: 21:42, Arrival Time: 23:26
Breakfast: Subway, Charlotte
Attraction: Books Monument, Charlotte.
Lunch: Olive Tree Cafe, Charlotte
Dinner: Kylin Skybar, Charlotte
Accommodation: -

***** Example Ends *****

Given information: {text}
Query: {query}
Travel Plan:
\end{lstlisting}

\subsubsection{React \& Reflexion Planning  Prompt}
The instruction prompts for the React and Reflexion planning strategies are provided as follows:

\lstset{
    backgroundcolor=\color[RGB]{245,245,245},
    breaklines=true,
    basicstyle=\ttfamily\small,
    frame=trbl,
    frameround = tttt,
}\begin{lstlisting}
You are a proficient planner. Based on the provided information and query, please give me a detailed plan, including specifics such as flight numbers (e.g., F0123456), restaurant names, and hotel names. Note that all the information in your plan should be derived from the provided data. You must adhere to the format given in the example. Additionally, all details should align with common sense. Attraction visits and meals are expected to be diverse. The symbol '-' indicates that information is unnecessary. For example, in the provided sample, you do not need to plan after returning to the departure city. When you travel to two cities in one day, you should note it in the 'Current City' section as in the example (i.e., from A to B). Solve this task by alternating between Thought, Action, and Observation steps. The 'Thought' phase involves reasoning about the current situation. 
The 'Action' phase can be of two types:
(1) CostEnquiry[Sub Plan]: This function calculates the cost of a detailed sub plan, which you need to input the people number and plan in JSON format. The sub plan should encompass a complete one-day plan. An example will be provided for reference.
(2) Finish[Final Plan]: Use this function to indicate the completion of the task. 
You must submit a final, complete plan as an argument.

***** Example *****
Query: Could you create a challenging travel plan for 7 people from Roanoke to Illinois spanning a week, from March 8th to March 14th, 2022, with a budget of $30,200? The preference is for an entire room, and we would not be taking any flights. In terms of cuisine, we are interested in sampling some Italian and Chinese food.You can call CostEuquiry like CostEuquiry[{{"people_number": 7,"day": 1,"current_city": "from Ithaca to Charlotte","transportation": "Flight Number: F3633413, from Ithaca to Charlotte, Departure Time: 05:38, Arrival Time: 07:46","breakfast": "Nagaland's Kitchen, Charlotte","attraction": "The Charlotte Museum of History, Charlotte","lunch": "Cafe Maple Street, Charlotte","dinner": "Bombay Vada Pav, Charlotte","accommodation": "Affordable Spacious Refurbished Room in Bushwick!, Charlotte"}}]
You can call Finish like Finish[
Day: 1
Current City: from Ithaca to Charlotte
Transportation: Flight Number: F3633413, from Ithaca to Charlotte, Departure Time: 05:38, 
Arrival Time: 07:46
Breakfast: Nagaland's Kitchen, Charlotte
Attraction: The Charlotte Museum of History, Charlotte
Lunch: Cafe Maple Street, Charlotte
Dinner: Bombay Vada Pav, Charlotte
Accommodation: Affordable Spacious Refurbished Room in Bushwick!, Charlotte

Day 2:
Current City: Charlotte
Transportation: -
Breakfast: Olive Tree Cafe, Charlotte
Attraction: The Mint Museum, Charlotte;Romare Bearden Park, Charlotte.
Lunch: Birbal Ji Dhaba, Charlotte
Dinner: Pind Balluchi, Charlotte
Accommodation: Affordable Spacious Refurbished Room in Bushwick!, Charlotte

Day 3:
Current City: Charlotte
Transportation: Flight Number: F3786167, from Charlotte to Ithaca, Departure Time: 21:42, Arrival Time: 23:26
Breakfast: Subway, Charlotte
Attraction: Books Monument, Charlotte.
Lunch: Olive Tree Cafe, Charlotte
Dinner: Kylin Skybar, Charlotte
Accommodation: -]
***** Example Ends *****

You must use Finish to indict you have finished the task. And each action only calls one function once.
Given information: {text}
Query: {query}
\end{lstlisting}

\subsubsection{Query Generation  Prompt}
The instruction prompt for query generation is provided as follows:
\label{appendix:prompt-query-generation}
\lstset{
    backgroundcolor=\color[RGB]{245,245,245},
    breaklines=true,
    basicstyle=\ttfamily\small,
    frame=trbl,
    frameround = tttt,
}\begin{lstlisting}
Given a JSON, please help me generate a natural language query. In the JSON, 'org' denotes the departure city. When 'days' exceeds 3, 'visiting_city_number' specifies the number of cities to be covered in the destination state. Here are three examples.

-----EXAMPLE 1-----
JSON:
{"org": "Gulfport", "dest": "Charlotte", "days": 3, "visiting_city_number": 1, "date": ["2022-03-05", "2022-03-06", "2022-03-07"], "people_number": 1, "constraint": {"room rule": null, "cuisine": null, "room type": null}, "budget": 1800}
QUERY:
Please design a travel plan departing Gulfport and heading to Charlotte for 3 days, spanning March 5th to March 7th, 2022, with a budget of $1800.
-----EXAMPLE 2-----
JSON:
{"org": "Omaha", "dest": "Colorado", "days": 5, "visiting_city_number": 2, "date": ["2022-03-14", "2022-03-15", "2022-03-16", "2022-03-17", "2022-03-18"], "people_number": 7, "constraint": {"room rule": "pets", "cuisine": null, "room type": null}, "budget": 35300}
QUERY:
Could you provide a  5-day travel itinerary for a group of 7, starting in Omaha and exploring 2 cities in Colorado between March 14th and March 18th, 2022? Our budget is set at $35,300, and it's essential that our accommodations be pet-friendly since we're bringing our pets.
-----EXAMPLE 3-----
JSON:
{"org": "Indianapolis", "dest": "Georgia", "days": 7, "visiting_city_number": 3, "date": ["2022-03-01", "2022-03-02", "2022-03-03", "2022-03-04", "2022-03-05", "2022-03-06", "2022-03-07"], "people_number": 2, "constraint": {"room rule": null, "cuisine": ["Bakery", "Indian"], "room type": "entire room", "transportation": "self driving"}, "budget": 6200}
QUERY:
I'm looking for a week-long travel itinerary for 2 individuals. Our journey starts in Indianapolis, and we intend to explore 3 distinct cities in Georgia from March 1st to March 7th, 2022. Our budget is capped at $6,200. For our accommodations, we'd prefer an entire room. We plan to navigate our journey via self-driving. In terms of food, we're enthusiasts of bakery items, and we'd also appreciate indulging in genuine Indian cuisine.
-----EXAMPLES END-----

JSON: {json}
QUERY:
\end{lstlisting}

\subsubsection{Key Components Extraction  Prompt}
The instruction prompt for plan key components extraction is provided as follows:
\label{appendx:prompt-extraction}
\lstset{
    backgroundcolor=\color[RGB]{245,245,245},
    breaklines=true,
    basicstyle=\ttfamily\small,
    frame=trbl,
    frameround = tttt,
}\begin{lstlisting}
Please assist me in extracting valid information from a given natural language text and reconstructing it in JSON format, as demonstrated in the following example. Use a ';' to separate different attractions, with each attraction formatted as 'Name, City'. If there's information about transportation, ensure that the 'current_city' aligns with the destination mentioned in the transportation details (i.e., the current city should follow the format 'from A to B'). Also, ensure that all flight numbers and costs are followed by a colon (i.e., 'Flight Number:' and 'Cost:'), consistent with the provided example. Each item should include ['day', 'current_city', 'transportation', 'breakfast', 'attraction', 'lunch', 'dinner', 'accommodation']. Replace non-specific information like 'eat at home/on the road' with '-'. Additionally, delete any '$' symbols.
-----EXAMPLE-----
 [{{
        "days": 1,
        "current_city": "from Dallas to Peoria",
        "transportation": "Flight Number: 4044830, from Dallas to Peoria, Departure Time: 13:10, Arrival Time: 15:01",
        "breakfast": "-",
        "attraction": "Peoria Historical Society, Peoria;Peoria Holocaust Memorial, Peoria;",
        "lunch": "-",
        "dinner": "Tandoor Ka Zaika, Peoria",
        "accommodation": "Bushwick Music Mansion, Peoria"
    }},
    {{
        "days": 2,
        "current_city": "Peoria",
        "transportation": "-",
        "breakfast": "Tandoor Ka Zaika, Peoria",
        "attraction": "Peoria Riverfront Park, Peoria;The Peoria PlayHouse, Peoria;Glen Oak Park, Peoria;",
        "lunch": "Cafe Hashtag LoL, Peoria",
        "dinner": "The Curzon Room - Maidens Hotel, Peoria",
        "accommodation": "Bushwick Music Mansion, Peoria"
    }},
    {{
        "days": 3,
        "current_city": "from Peoria to Dallas",
        "transportation": "Flight Number: 4045904, from Peoria to Dallas, Departure Time: 07:09, Arrival Time: 09:20",
        "breakfast": "-",
        "attraction": "-",
        "lunch": "-",
        "dinner": "-",
        "accommodation": "-"
    }}]
-----EXAMPLE ENDS-----
Text: {text}
JSON: 
\end{lstlisting}

\section{Case Presentation}
\label{appendix:case}

\subsection{Example of Query and Reference Plan}
\label{appendix:case-query-plan}
we present an example of a query and its corresponding reference plan in our train set as follows:
\lstset{
    backgroundcolor=\color[RGB]{245,245,245},
    breaklines=true,
    basicstyle=\ttfamily\small,
    frame=trbl,
    frameround = tttt,
}\begin{lstlisting}
{
    "org": "Indianapolis",
    "dest": "Colorado",
    "days": 7,
    "visiting_city_number": 3,
    "date": [
        "2022-03-11",
        "2022-03-12",
        "2022-03-13",
        "2022-03-14",
        "2022-03-15",
        "2022-03-16",
        "2022-03-17"
    ],
    "people_number": 5,
    "room rule": "pets",
    "cuisine": [
        "Mexican",
        "Italian",
        "Mediterranean",
        "Indian"
    ],
    "room type": "entire room",
    "transportation": null,
    "budget": 15100,
    "query": "Can you help with generating a 7-day travel plan for a party of 5? We're setting off from Indianapolis and planning to explore 3 cities in Colorado from March 11th to March 17th, 2022. We have a budget of $15,100 for this trip. We'll be bringing our pets, so pet-friendly accommodations are a must. We're also hoping to find places that offer Mexican, Italian, Mediterranean, and Indian cuisines. Entire rooms for accommodations would be ideal.",
    "level": "hard",
    "annotated plan": [
        {
            "days": 1,
            "current_city": "from Indianapolis to Grand Junction(Colorado)",
            "transportation": "Self-driving, from Indianapolis to Grand Junction(Colorado), duration: 19 hours 21 mins, distance: 2,132 km, cost: 106",
            "breakfast": "-",
            "attraction": "-",
            "lunch": "-",
            "dinner": "Nukkadwala, Grand Junction(Colorado)",
            "accommodation": "Lovely 1 BD on the Upper West Side, Grand Junction(Colorado)"
        },
        {
            "days": 2,
            "current_city": "Grand Junction(Colorado)",
            "transportation": "-",
            "breakfast": "Om Ji Bhature Wale, Grand Junction(Colorado)",
            "attraction": "Museum of the West, Museums of Western Colorado, Grand Junction(Colorado);Eureka! McConnell Science Museum, Grand Junction(Colorado);",
            "lunch": "Penta Cafe, Grand Junction(Colorado)",
            "dinner": "Kings Kulfi, Grand Junction(Colorado)",
            "accommodation": "Lovely 1 BD on the Upper West Side, Grand Junction(Colorado)"
        },
        {
            "days": 3,
            "current_city": "from Grand Junction(Colorado) to Alamosa(Colorado)",
            "transportation": "Self-driving, from Grand Junction(Colorado) to Alamosa(Colorado), duration: 4 hours 37 mins, distance: 397 km, cost: 19",
            "breakfast": "Punjab Da Pind, Grand Junction(Colorado)",
            "attraction": "Alamosa Colorado Welcome Center, Alamosa(Colorado);Toivo Malm Trail System, Alamosa(Colorado);",
            "lunch": "Emperor's Lounge - The Taj Mahal Hotel, Alamosa(Colorado)",
            "dinner": "Cafe Dalal Street, Alamosa(Colorado)",
            "accommodation": "Sunny Chelsea Studio, Alamosa(Colorado)"
        },
        {
            "days": 4,
            "current_city": "Alamosa(Colorado)",
            "transportation": "-",
            "breakfast": "Good Luck Cafe, Alamosa(Colorado)",
            "attraction": "Alamosa Archery Range, Alamosa(Colorado);Alamosa Riparian Park, Alamosa(Colorado);Alamosa Sub, Alamosa(Colorado);",
            "lunch": "Shri Durga Dosa Corner, Alamosa(Colorado)",
            "dinner": "Lahore, Alamosa(Colorado)",
            "accommodation": "Sunny Chelsea Studio, Alamosa(Colorado)"
        },
        {
            "days": 5,
            "current_city": "from Alamosa(Colorado) to Denver(Colorado)",
            "transportation": "Self-driving, from Alamosa(Colorado) to Denver(Colorado), duration: 3 hours 38 mins, distance: 377 km, cost: 18",
            "breakfast": "Hamburg To Hyderabad, Alamosa(Colorado)",
            "attraction": "Denver Zoo, Denver(Colorado);",
            "lunch": "The Fatty Bao - Asian Gastro Bar, Denver(Colorado)",
            "dinner": "Woods Spice, Denver(Colorado)",
            "accommodation": "Peaceful, beautiful home away , Denver(Colorado)"
        },
        {
            "days": 6,
            "current_city": "Denver(Colorado)",
            "transportation": "-",
            "breakfast": "The Urban Socialite, Denver(Colorado)",
            "attraction": "Denver Art Museum, Denver(Colorado);Denver Museum of Nature & Science, Denver(Colorado);",
            "lunch": "Breaktym, Denver(Colorado)",
            "dinner": "Chawla's\u5b8a, Denver(Colorado)",
            "accommodation": "Peaceful, beautiful home away , Denver(Colorado)"
        },
        {
            "days": 7,
            "current_city": "from Denver(Colorado) to Indianapolis",
            "transportation": "Self-driving, from Denver(Colorado) to Indianapolis, duration: 15 hours 37 mins, distance: 1,747 km, cost: 87",
            "breakfast": "Starve Stalkers, Denver(Colorado)",
            "attraction": "-",
            "lunch": "-",
            "dinner": "-",
            "accommodation": "-"
        }
    ]
}
\end{lstlisting}

\subsection{Additional Case Study}
\label{appendix:case-study}
\begin{figure*}[t]
    \centering
    \includegraphics[width=0.4\linewidth]{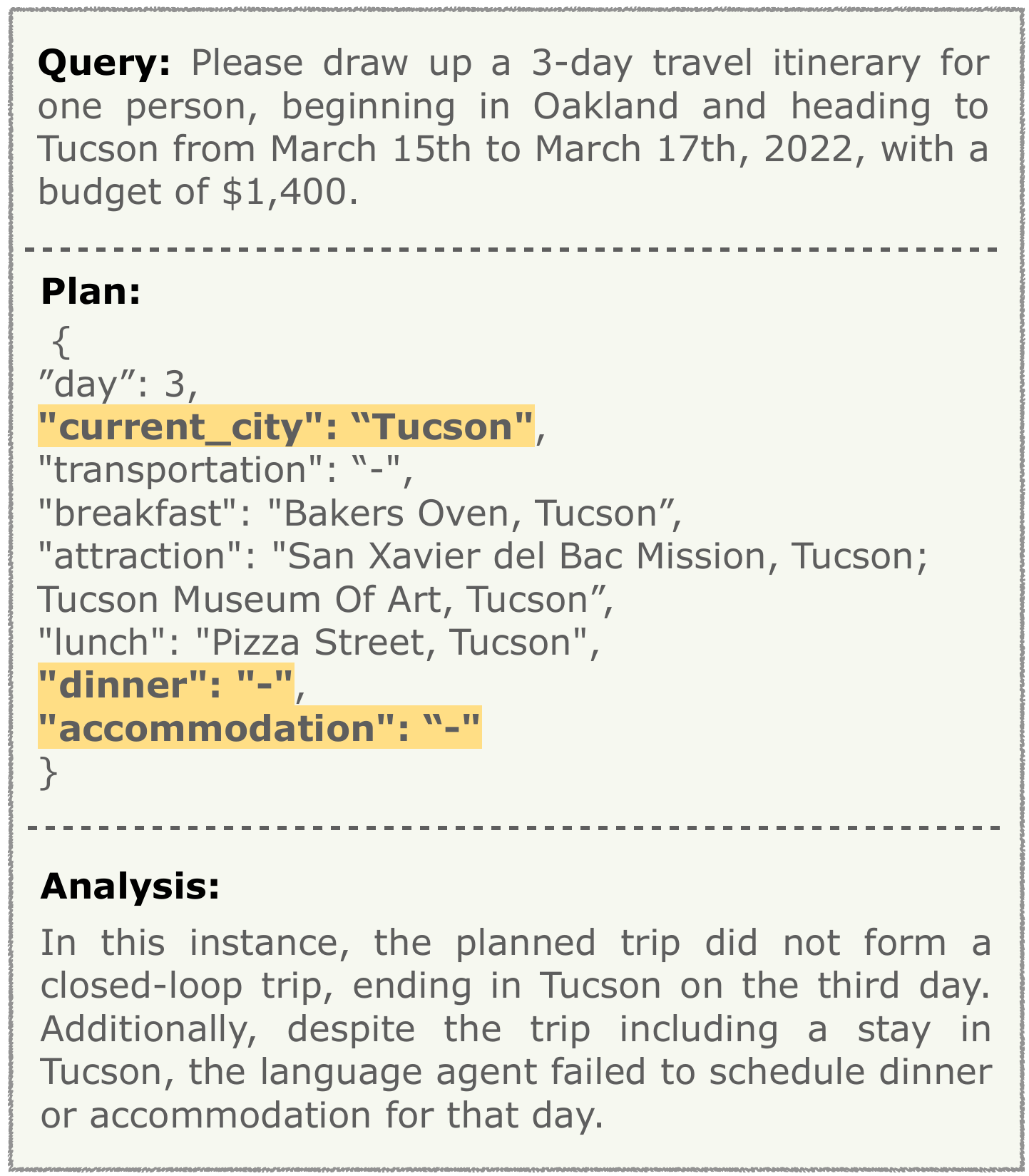}
    \caption{Case 4 (Two-stage Mode). }
\end{figure*}

\begin{figure*}[t]
    \centering
    \includegraphics[width=0.4\linewidth]{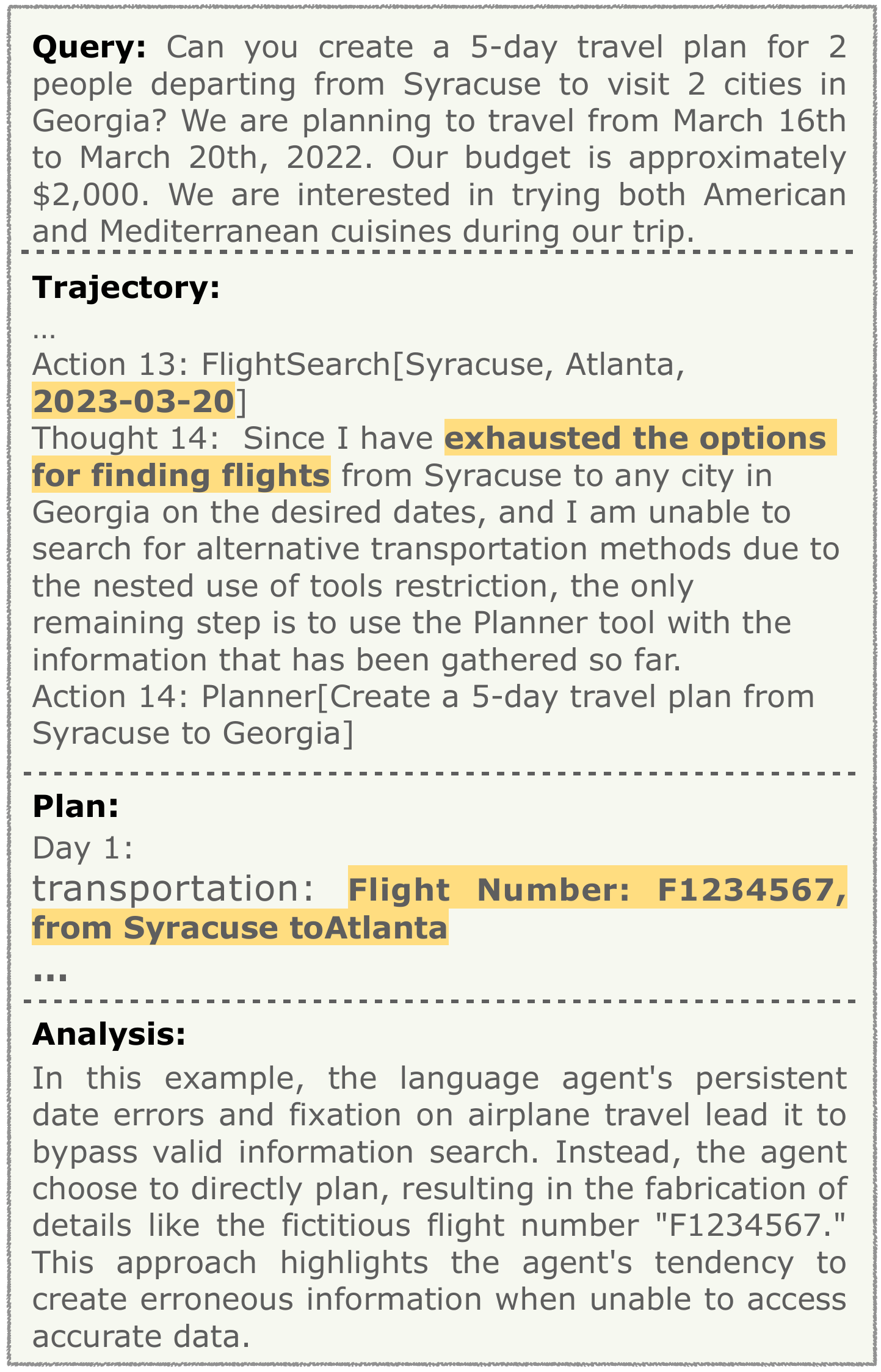}
    \caption{Case 5 (Two-stage mode).}
\end{figure*}

\begin{figure*}[t]
    \centering
    \includegraphics[width=0.4\linewidth]{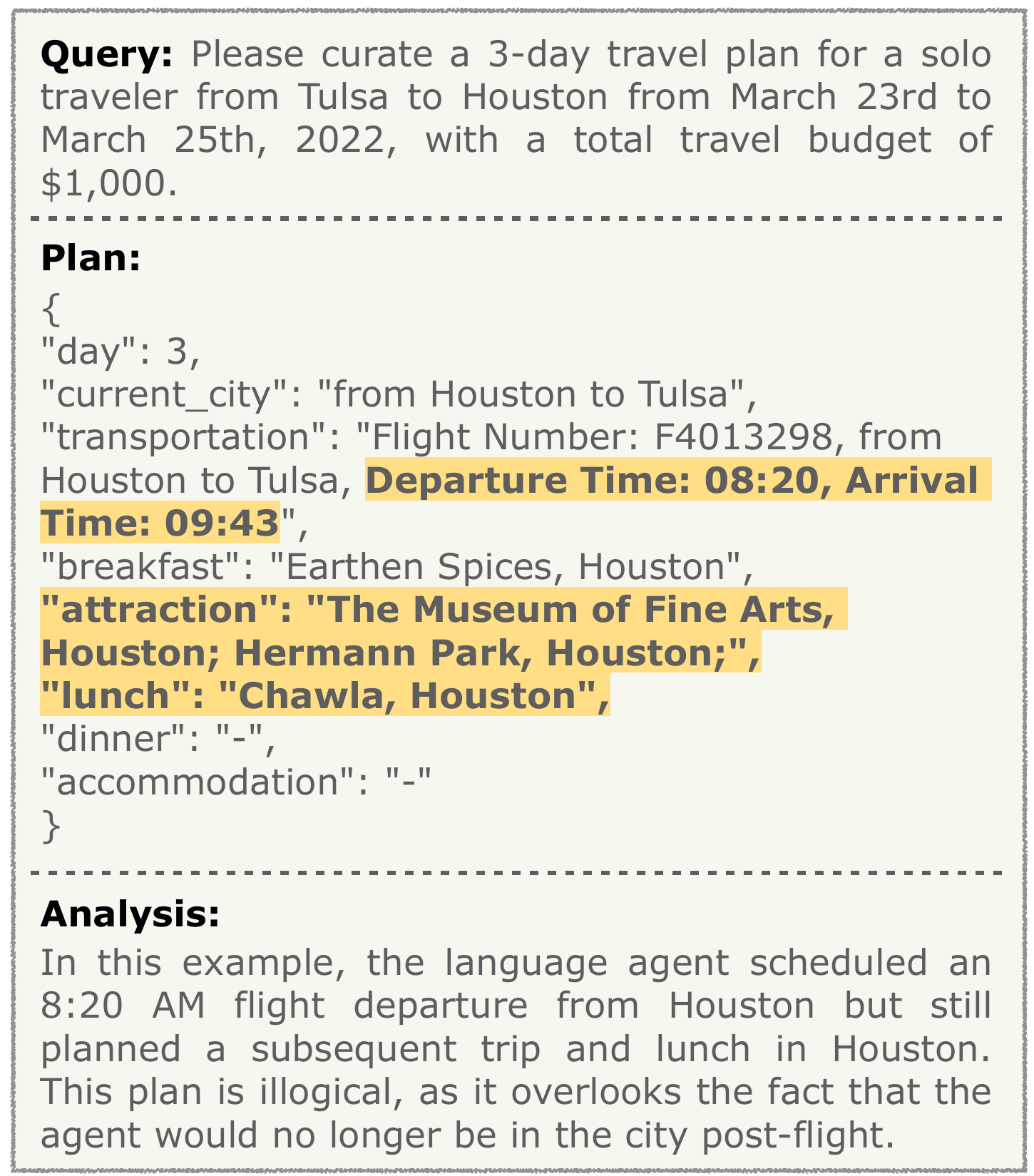}
    \caption{Case 6 (Two-stage mode).}
\end{figure*}

\begin{figure*}[t]
    \centering
    \includegraphics[width=0.4\linewidth]{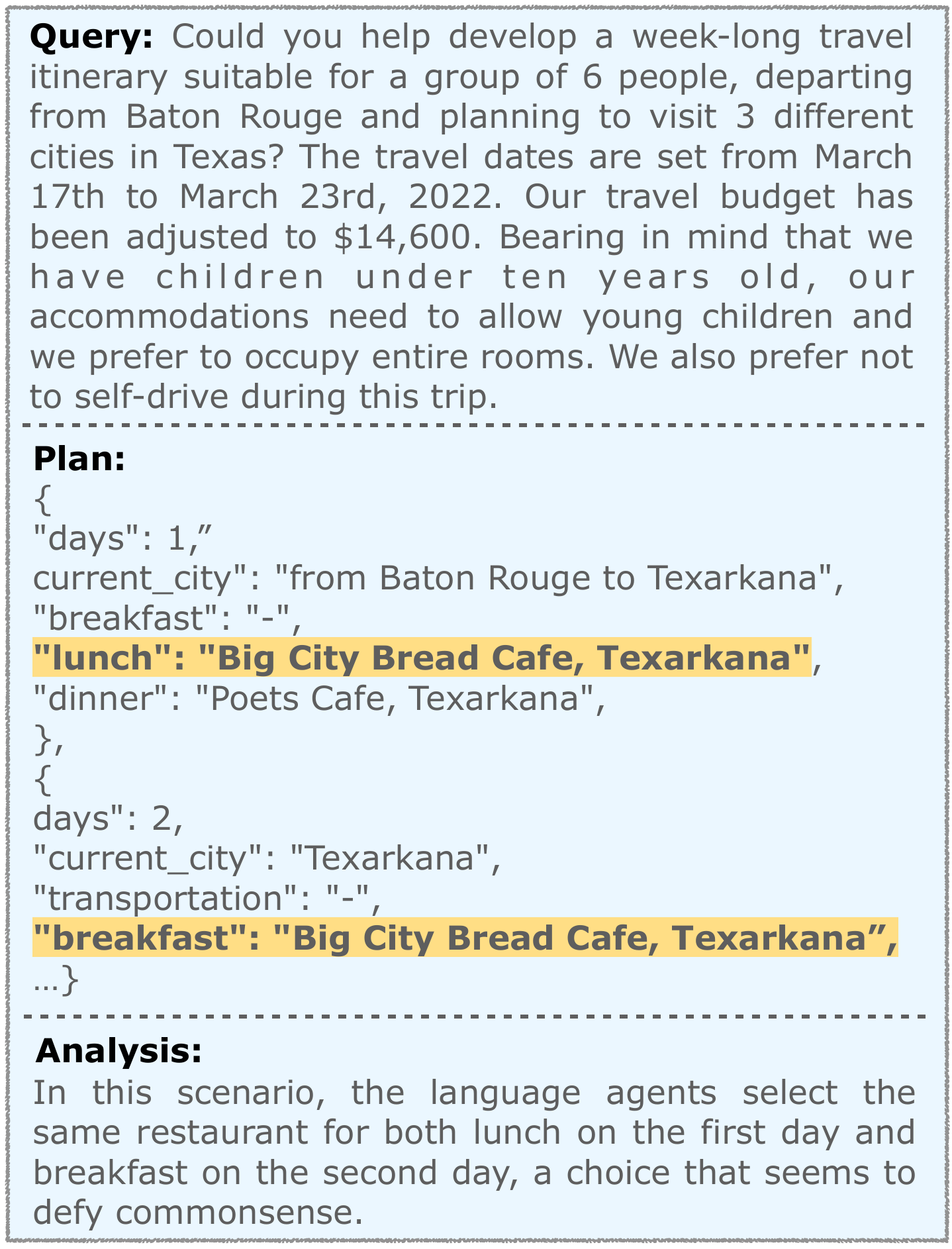}
    \caption{Case 7 (Direct strategy in sole-planning mode).}
\end{figure*}

\begin{figure*}[t]
    \centering
    \includegraphics[width=0.4\linewidth]{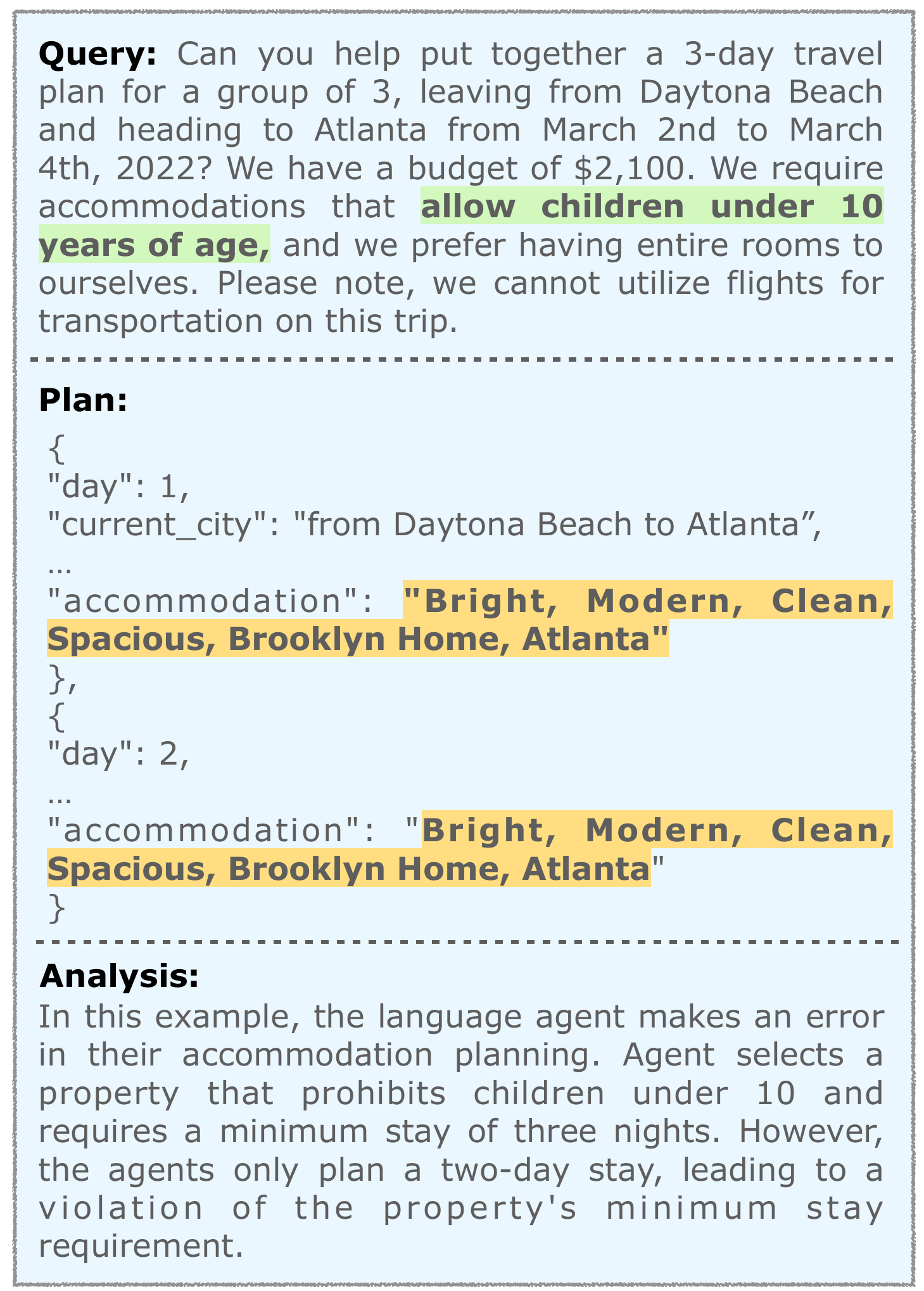}
    \caption{Case 8 (Direct strategy in sole-planning mode).}
\end{figure*}

\begin{figure*}[t]
    \centering
    \includegraphics[width=0.4\linewidth]{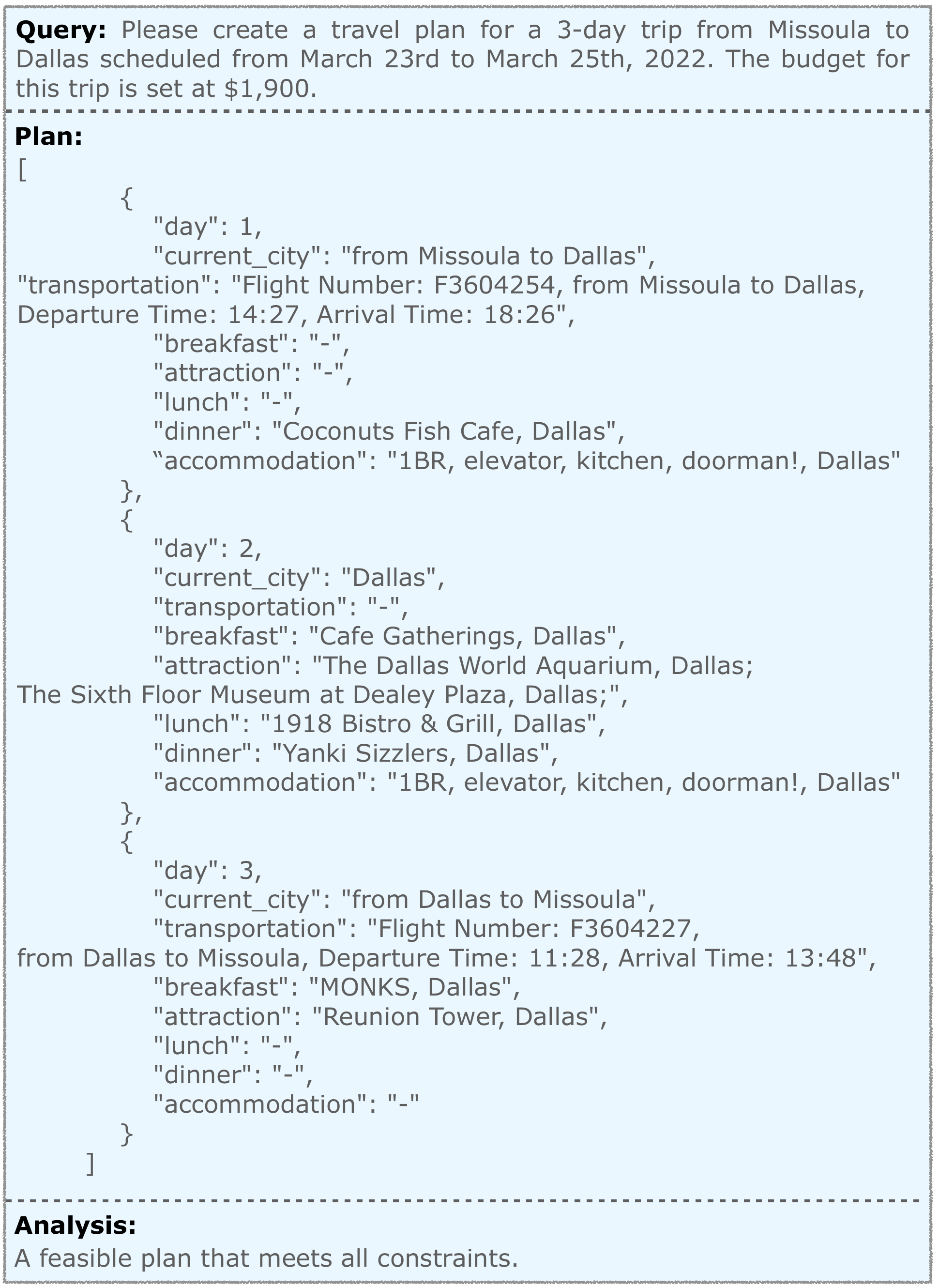}
    \caption{Case 9 (Direct strategy in sole-planning mode).}
\end{figure*}

\begin{figure*}[t]
    \centering
    \includegraphics[width=0.5\linewidth]{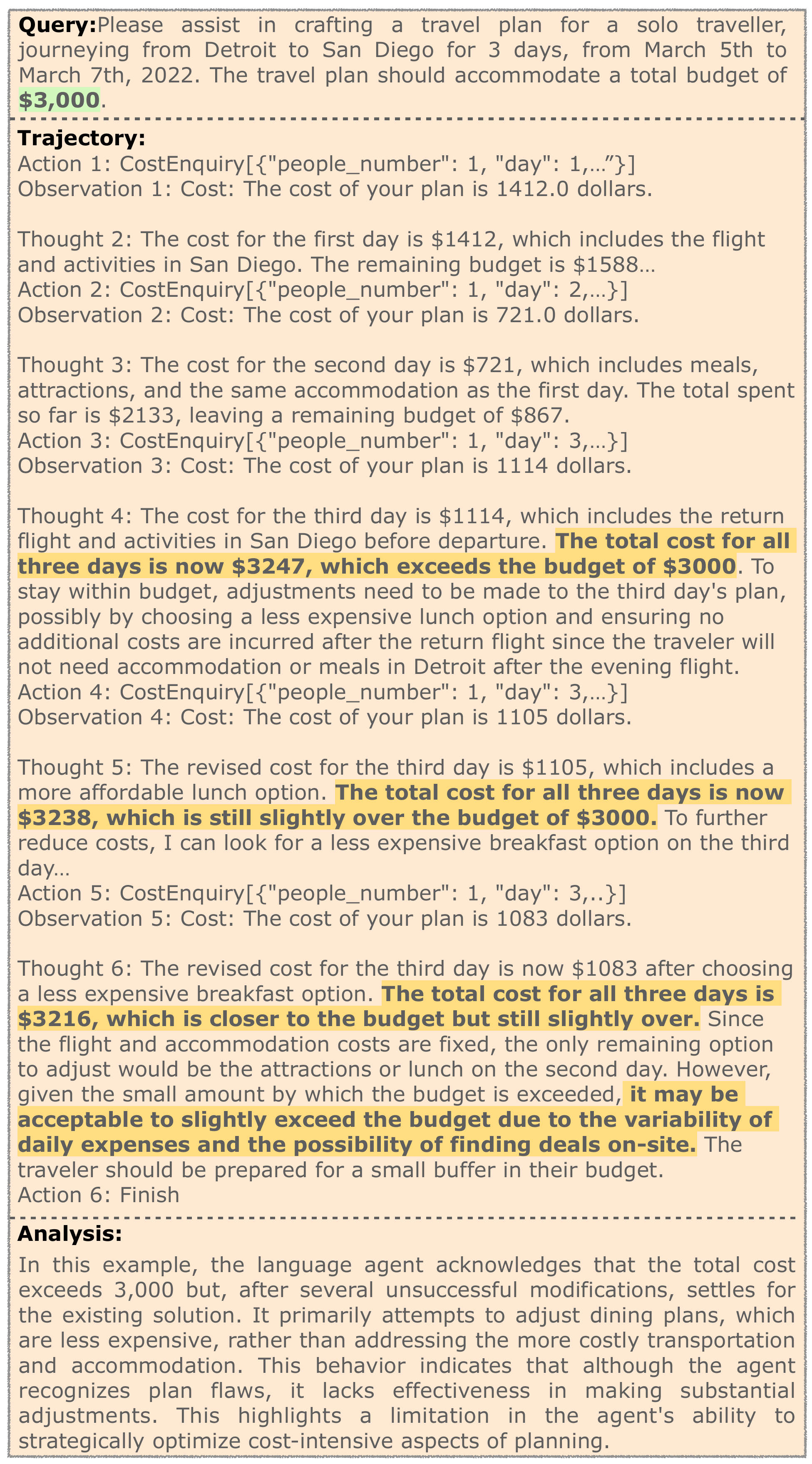}
    \caption{Case 10 (Reflexion strategy in sole-planning mode).}
\end{figure*}

%% file: tables/dataset_distribution.tex
\resizebox{0.5\linewidth}{!}{
\begin{tabular}{lcccc}
\toprule
                                     & Days  & Easy & Medium & Hard \\ \midrule
\multirow{3}{*}{Training (\#\num{45})}        & 3-day & \num{5}    & \num{5}      & \num{5}    \\
                                     & 5-day & \num{5}    & \num{5}      & \num{5}    \\
                                     & 7-day & \num{5}    & \num{5}      & \num{5}    \\ \midrule
\multirow{3}{*}{Validation (\#\num{180})} & 3-day & \num{20}   & \num{20}     & \num{20}   \\
                                     & 5-day & \num{20}   & \num{20}     & \num{20}   \\
                                     & 7-day & \num{20}   & \num{20}     & \num{20}   \\ \midrule
\multirow{3}{*}{Test (\#\num{1000})}       & 3-day & \num{122}  & \num{104}    & \num{82}   \\
                                     & 5-day & \num{116}  & \num{114}    & \num{121}  \\
                                     & 7-day & \num{110}  & \num{115}    & \num{116}  \\ \bottomrule
\end{tabular}}

%% file: tables/tool_description.tex
\resizebox{\linewidth}{!}{
\begin{tabular}{lll}
\toprule
Tool                 & Data Entries(\#) & Description \\ \hline
CitySearch           & \num{312}       & \parbox{0.75\linewidth}{Search cities in the given state.} \\
FlightSearch         & \num{3827361} & \parbox{0.75\linewidth}{Search flight information for a specific date between two cities.} \\
DistanceMatrix & \num{17603}    & \parbox{0.75\linewidth}{Search the driving distance, time, and possible cost between two cities.} \\
RestaurantSearch     & \num{9552}     & \parbox{0.75\linewidth}{Search restaurants in the given city.} \\
AttractionSearch     & \num{5303}     & \parbox{0.75\linewidth}{Search attractions in the given city.} \\
AccommodationSearch & \num{5064}     & \parbox{0.75\linewidth}{Search accommodations in the given city.} \\
NotebookWrite        & -         & \parbox{0.75\linewidth}{Write the selected data entry into the Notebook tool with a short description. 
} \\
\bottomrule
\end{tabular}
}

%% file: tables/gpt4_relfexion.tex
\begin{tabular}{lcccccc}
\toprule
                           & \multirow{2}{*}{\begin{tabular}[c]{@{}c@{}}Delivery\\ Rate\end{tabular}} & \multicolumn{2}{c}{\begin{tabular}[c]{@{}c@{}}Commonsense\\ Pass Rate\end{tabular}} & \multicolumn{2}{c}{\begin{tabular}[c]{@{}c@{}}Hard Constraint \\ Pass Rate\end{tabular}} & \multirow{2}{*}{\begin{tabular}[c]{@{}c@{}}Final \\ Pass Rate\end{tabular}} \\ \cmidrule(l){3-4} \cmidrule(l){5-6}
                           &                                                                          & Micro                                    & Macro                                    & Micro                                       & Macro                                      &                                                                             \\ \midrule
Reflexion$_{\rm GPT-4-Turbo}$ & \num{80.6}                                                                     & \num{62.9}                                     & \num{6.1}                                      & \num{52.4}                                        & \num{40.0}                                       & \num{3.3}                                                                           \\ \bottomrule
\end{tabular}